\documentclass[fleqn,10pt]{wlscirep}
\usepackage[utf8]{inputenc}
\usepackage[T1]{fontenc}
\usepackage{graphicx}
\usepackage{authblk}
\usepackage{siunitx}
\usepackage{booktabs}
\usepackage{caption}
\usepackage{float}
\usepackage{color,soul}
\usepackage{bm}
\usepackage{makecell}
\usepackage{xurl}
\usepackage{tabularx}   
\usepackage{multirow}
\usepackage{times}
\usepackage[title]{appendix}
\usepackage{amsmath}
\usepackage{amsfonts}
\usepackage{amssymb}
\usepackage{xcolor}
\usepackage{xr}
\usepackage{multirow}
\usepackage{booktabs}
\usepackage[table]{xcolor}
\usepackage{threeparttable}
\usepackage{graphicx}
\usepackage{booktabs,longtable,pdflscape,multirow,threeparttable,siunitx}
\newcommand{\pier}{\textsc{Pier}}
\newcommand{\IQL}{\textsc{Iql}}
\newcommand{\COtwo}{CO\textsubscript{2}}
\newcommand{\Hs}{H_s}
\newcommand{\Tp}{T_p}

\hypersetup{
    colorlinks=true,
    linkcolor=blue!60!black,
    citecolor=blue!60!black,
    urlcolor=blue!60!black,
}

\usepackage{array}
\usepackage[super,sort&compress,numbers]{natbib}
\usepackage{hyperref}

\makeatletter

\newcommand*{\addFileDependency}[1]{
\typeout{(#1)}
\@addtofilelist{#1}
\IfFileExists{#1}{}{\typeout{No file #1.}}
}\makeatother

\newcommand*{\myexternaldocument}[1]{%
\externaldocument{#1}%
\addFileDependency{#1.tex}%
\addFileDependency{#1.aux}%
}

\myexternaldocument{supplementary}
\title{Physics-informed offline reinforcement learning eliminates catastrophic fuel waste in maritime routing}

\author[1$\ast$]{Aniruddha Bora}
\author[2]{Julie Chalfant}
\author[2]{Chryssostomos Chryssostomidis}

\affil[1]{Department of Computer Science, Texas State University, San Marcos, TX 78666, USA}
\affil[2]{MIT Sea Grant College Program, Massachusetts Institute of Technology, Cambridge, MA 02139, USA}
\affil[$\ast$]{Corresponding author}
\begin{abstract}
International shipping produces approximately 3\% of global greenhouse gas emissions, yet voyage routing remains dominated by heuristic methods that cannot adapt to spatiotemporally varying ocean conditions. Here we present \pier{} (Physics-Informed, Energy-efficient, Risk-aware routing), an offline reinforcement learning framework that learns fuel-efficient, safety-aware routing policies from physics-calibrated environments grounded in historical vessel tracking data and ocean reanalysis products, requiring no online simulator or interaction. Validated on one full year (2023) of Automatic Identification System data across seven Gulf of Mexico routes comprising 840 simulated evaluation episodes per method, \pier{} reduces estimated mean \COtwo{} emissions by 10\% relative to great-circle routing in simulation. However, the mean masks the framework's primary contribution: eliminating catastrophic fuel waste. Under typical conditions, median savings are 1\%, but \pier{} reduces 95th-percentile \COtwo{} by 6\% and maximum single-voyage emissions by 70\%. Great-circle routing incurs extreme fuel consumption ($>$1.5$\times$ median) in 4.8\% of voyages; \pier{} reduces this to 0.5\%- a 9-fold reduction in worst-case frequency. Per-voyage fuel consumption variance is 3.5$\times$ lower with \pier{} than great-circle routing ($p < 0.001$). For fleet operators, this predictability translates to reliable emission budgeting across Gulf routes (bootstrap 95\% CI for mean savings: [2.9\%, 15.7\%]). Partial validation against observed AIS vessel behavior on a cross-Gulf corridor confirms that \pier{}'s predictions are consistent with the fastest real transits while exhibiting $23.1\times$ lower variance. Crucially, \pier{}'s performance is forecast-independent: unlike classical path optimization whose wave protection degrades $4.5\times$ under realistic forecast uncertainty, \pier{} maintains constant performance using only local observations. More broadly, \pier{} demonstrates that physics-informed offline reinforcement learning can provide operationally reliable decision-making in safety-critical transportation domains where simulators are unavailable.
\end{abstract}

\begin{document}
\flushbottom
\maketitle

\thispagestyle{empty}

\setcounter{secnumdepth}{0}

\section*{Introduction}

Maritime shipping is the circulatory system of the global economy, transporting over 80\% of world trade by volume~\citep{unctad2023}. It is also a major source of greenhouse gas emissions, and a key target for machine learning-driven decarbonisation~\citep{rolnick2022tackling}. the International Maritime Organization (IMO) estimates that international shipping produced 1,076 million tonnes of \COtwo{} in 2018, approximately 2.9\% of global anthropogenic emissions~\citep{imo2020ghg}. Under the IMO's revised greenhouse gas strategy, the sector must achieve at least 40\% reduction in carbon intensity by 2030 and net-zero emissions by or around 2050~\citep{imo2023strategy}. Weather routing, or adjusting vessel speed and heading based on ocean conditions, is the lowest-cost decarbonization lever available today, requiring no hardware retrofit and offering immediate operational savings~\citep{psaraftis2019speed}.

Yet the current state of practice lags far behind what is technically possible. Most commercial weather routing tools rely on isochrone methods or parametric optimization that treat the problem as deterministic~\citep{mannarini2016graph,zis2020ship, walther2016modeling,szlapczynska2015multi,zhang2007optimal,simonsen2015state}. They do not learn from historical outcomes, cannot adapt to distributional shifts in weather patterns, and provide single-point forecasts without risk quantification. Academic approaches using reinforcement learning (RL) have shown 
promise in atmospheric navigation~\citep{bellemare2020autonomous} and maritime settings~\citep{wang2020rl_shipping,latinopoulos2025marine, zhao2025intelligent}, but 
require online simulators that do not exist for realistic ocean environments, the coupled dynamics of waves, currents, wind, vessel hull response, and engine performance are too complex and too poorly constrained for faithful simulation.

The maritime domain offers a unique opportunity to combine 
physics-informed machine learning~\citep{karniadakis2021physics, kadambi2023incorporating, li2024physics,toscano2025pinns} with offline reinforcement learning~\citep{levine2020offline}. Millions of historical voyages are recorded via the Automatic Identification System (AIS), paired with high-resolution ocean reanalysis data from products such as the Copernicus Marine Service~\citep{copernicus2023waves} and NOAA CoastWatch~\citep{pichel2000noaa}. Together, these enable calibration of physics-based environment models from which diverse training trajectories can be generated - a form of offline RL where the ``offline dataset'' is not raw logged behavior but trajectories sampled from an AIS-calibrated environment. This approach requires solving three domain-specific challenges: (1) raw AIS data lacks the physics-informed features needed for meaningful state representations; (2) generated trajectories must be behaviorally diverse to ensure broad state-action coverage; and (3) routing decisions are safety-critical no algorithm should route a vessel through dangerous seas. While prior work has fused AIS with meteorological data for fuel consumption prediction~\citep{du2022fusion,ren2022carbon}, these approaches typically match voyage-level aggregates to environmental averages, discarding the sub-hourly spatiotemporal structure needed for routing decisions.

We address all three challenges with \pier{}, a unified framework comprising three components. First, \emph{physics-informed state construction}: we fuse AIS vessel kinematics with wave, wind, and current reanalysis data to calibrate a speed-loss model and construct operationally meaningful state features including speed-loss estimates, hull-fatigue risk exposure, and energy expenditure indicators. Second, \emph{demonstration-augmented offline datasets}: we generate training data within the AIS-calibrated environment by mixing A*-optimal teacher trajectories that encode domain knowledge with stochastic behavioral roll-outs that provide broad state-action coverage. Third, a \emph{decoupled post-hoc safety shield}: rather than embedding safety constraints in the reward function (which distorts learning) \citep{garcia2015comprehensive}, we enforce hard navigational safety constraints at evaluation time through a computationally lightweight shield that prevents land collisions and hazardous wave exposure \citep{brunke2022safe}.

We validate \pier{} on a full year (2023) of operational AIS data across seven Gulf of Mexico routes, spanning all four meteorological seasons including hurricane season. Our central finding is twofold. First, \pier{}'s primary value is not average fuel savings but \emph{variance reduction}: it eliminates the catastrophic fuel-wasting events that great-circle routing produces 4.8\% of the time. Second, unlike classical path optimization (A*), \pier{}'s performance is \emph{forecast-independent}: A*'s wave protection degrades $4.5\times$ under realistic forecast uncertainty, while \pier{} maintains constant performance using only local observations - a decisive advantage for operational deployment where forecast quality beyond 48 hours is unreliable.

Beyond maritime routing, \pier{} establishes a general recipe for deploying offline RL in safety-critical physical domains that lack simulators: use domain physics to construct informative state features from sensor data, augment expert demonstrations with diverse behavioral roll-outs in the calibrated environment, and enforce safety through post-hoc constraints. This pattern transfers to wildfire evacuation routing, aircraft trajectory optimization, and autonomous navigation in unmapped terrain.

\section*{Results}

\subsection*{Physics-informed features capture operationally relevant variation}

The foundation of \pier{} is a state representation that encodes the physical interaction between vessel and environment. Rather than using raw AIS features (position, speed over ground, heading), we compute physics-informed features that capture the mechanisms by which ocean conditions affect vessel performance.

The speed-loss model, fitted to one full year of Gulf of Mexico AIS data matched with Copernicus wave reanalysis, takes the form:
\begin{equation}
    \Delta U = a \cdot \frac{\Hs}{\Tp} \cdot \cos(\mu)^{1.5} + b \cdot \Hs^2 + c \cdot C_{\text{tail}} + d \cdot V_{\text{tail}} + e
    \label{eq:speed_loss}
\end{equation}
where $\Hs$ is significant wave height, $\Tp$ is peak wave period, $\mu$ is the relative wave encounter angle, $C_{\text{tail}}$ is the along-track current component, and $V_{\text{tail}}$ is the along-track wind component. The fitted coefficients for cargo vessels are $a = -0.377$, $b = 0.002$, $c = 0.837$, $d = 0.029$, and $e = 0.139$ ($R^2 = 0.020$, $N = 8{,}149$ transits).

The modest $R^2$ is expected and physically appropriate: a vessel's  speed is predominantly determined by operational decisions (throttle setting, schedule adherence, draft 
configuration)~\citep{bialystocki2016estimation,adland2020optimal} and only marginally by environmental forcing. . The critical question is not whether the model explains total speed variance, but whether it correctly captures the \emph{directional dependence} of environmental effects and binned analysis confirms that it does, with speed loss increasing monotonically with wave height and steepness (Extended Data Fig.~1). In a controlled simulated environment where ground-truth coefficients are known, the regression pipeline recovers the exact speed-loss model despite comparable $R^2$ levels, confirming that low $R^2$ reflects irreducible operational noise rather than model failure (Supplementary Section~1.2).

\begin{figure}[t]
\centering
\includegraphics[width=0.65\textwidth]{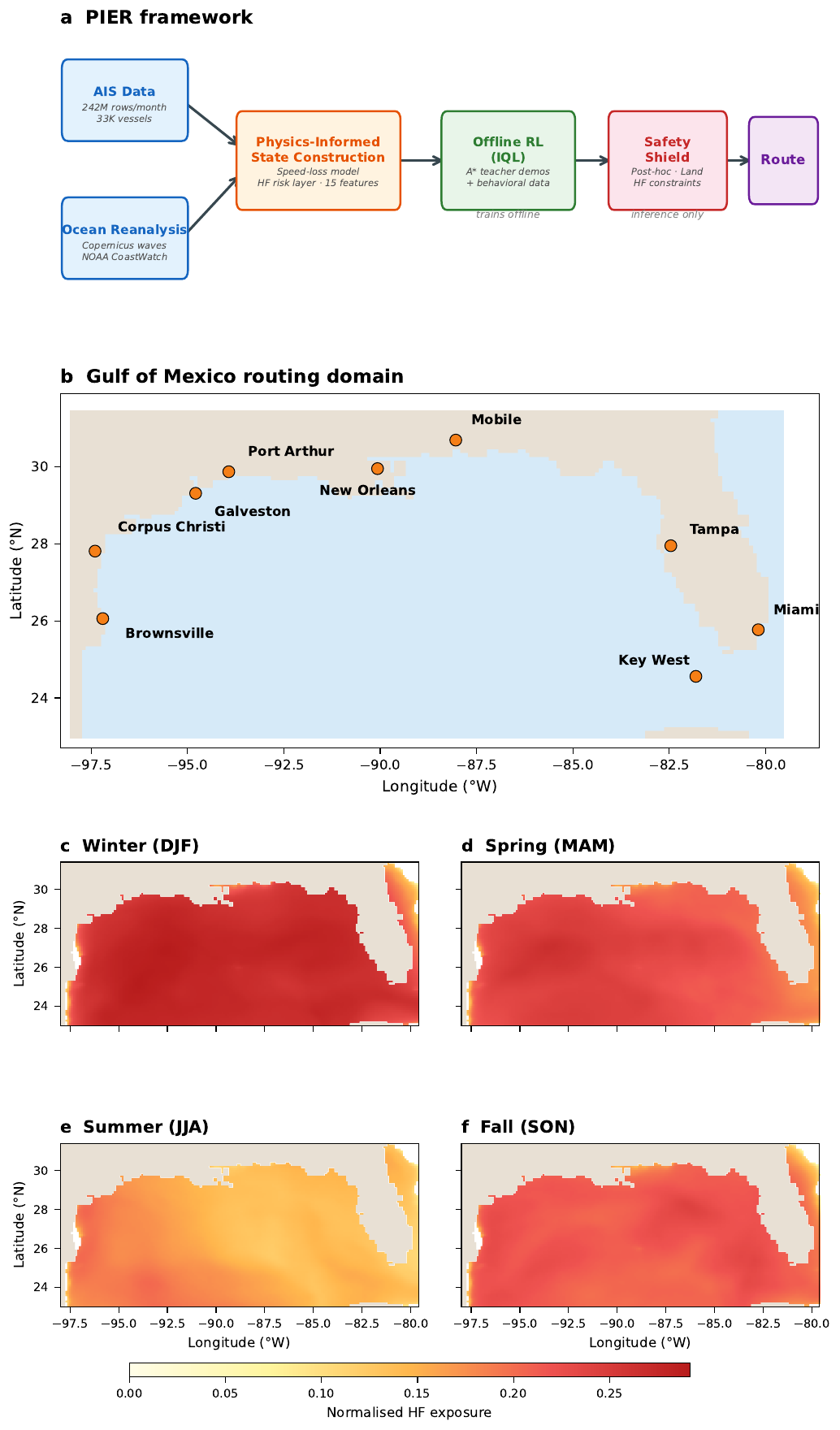}
\caption{\textbf{\pier{} framework and Gulf of Mexico routing domain.}
\textbf{a}, The \pier{} pipeline: AIS vessel tracking data and ocean reanalysis products (Copernicus Marine Service, NOAA CoastWatch ERDDAP) are fused into physics-informed state features (speed-loss model, hull-fatigue risk layer, 15 normalized features). An IQL agent is trained offline on a mixed dataset of A*-optimal teacher demonstrations and stochastic behavioral roll-outs. At inference, a post-hoc safety shield enforces land-avoidance and wave-exposure constraints before each routing decision.
\textbf{b}, Gulf of Mexico routing domain with nine key ports (amber dots). Seven port-to-port routes are evaluated: Brownsville$\to$Key West, Corpus Christi$\to$Galveston, Corpus Christi$\to$Tampa, Galveston$\to$Key West, Mobile$\to$Tampa, New Orleans$\to$Tampa, and Port Arthur$\to$Mobile.
\textbf{c--f}, Seasonal mean normalized HF (hull fatigue) exposure from Copernicus wave reanalysis, showing elevated hazard in winter (DJF) and fall (SON) compared with summer (JJA), particularly in the eastern Gulf and Florida Straits. This seasonal variation is the primary driver of \pier{}'s fuel savings: weather-aware routing adds the most value when conditions are adverse.}
\label{fig:seasonal}
\end{figure}

From the wave fields, we construct a Hull Fatigue (HF) risk layer that quantifies the cumulative wave exposure hazard at each grid cell:
\begin{equation}
    E_{\text{HF}} = \frac{\Hs}{\Tp} \cdot \max(0, \cos\mu)^{1.5}
    \label{eq:hf}
\end{equation}
This exposure metric varies dramatically across seasons (Fig.~\ref{fig:seasonal}c--f): winter (DJF) and fall (SON) show 2--3$\times$ higher HF exposure than summer (JJA), confirming that weather-aware routing matters most during adverse conditions.

\subsection*{\pier{} learns fuel-efficient routes with consistent performance across seasons}

We evaluate \pier{} (\IQL{} with physics-informed states and safety shield) against five baselines across 840 evaluation episodes spanning all 12 months of 2023 and seven Gulf of Mexico routes (Table~\ref{tab:baselines}). Online RL (DQN, 0.4\% arrival) and supervised learning (behavioural cloning, 
7.5\% arrival) fail completely in this offline setting, confirming the necessity of offline RL (Extended Data Fig. 2).  Among viable methods, \pier{} achieves 83.3\% arrival rate exceeding great-circle routing (78.0\%), while delivering 8\% faster mean transit time (45.6\,h versus 49.8\,h) and 5\% lower wave exposure (53.2 versus 55.8).  
Arrival rate indicates the percentage of simulations that reach the destination.

\begin{table}[t]
\centering
\caption{\textbf{Annual baseline comparison (2023, 12 months, all 7 routes).} Arrival rate, transit time, estimated fuel proxy, and HF wave exposure (mean $\pm$ std). Bold indicates best among methods with $>$50\% arrival rate.}
\label{tab:baselines}
\begin{tabular}{llcccc}
\toprule
\textbf{Method} & \textbf{Type} & \textbf{Arrival} & \textbf{Time (h)} & \textbf{Fuel Proxy} & \textbf{HF Exposure} \\
\midrule
Great Circle     & Heuristic   & 655/840 (78.0\%) & 49.8 $\pm$ 52.8  & 89.5 $\pm$ 94.4  & 55.8 $\pm$ 56.1 \\
Greedy Goal      & Heuristic   & 685/840 (81.5\%) & 46.6 $\pm$ 42.5  & 78.8 $\pm$ 74.9   & 51.7 $\pm$ 48.4 \\
BC (expert)      & Supervised  & 63/840 (7.5\%)   & 101.1 $\pm$ 60.5  & 117.3 $\pm$ 81.0   & 83.6 $\pm$ 62.7 \\
DQN              & Online RL   & 3/840 (0.4\%)    & 70.8 $\pm$ 140.0  & 79.5 $\pm$ 157.0   & 37.8 $\pm$ 75.8  \\
CQL              & Offline RL  & 643/840 (76.5\%) & 49.7 $\pm$ 35.0  & 83.9 $\pm$ 60.9   & 55.3 $\pm$ 46.5 \\
\textbf{\pier{} (IQL)} & \textbf{Offline RL} & \textbf{700/840 (83.3\%)} & \textbf{45.6 $\pm$ 26.4} & \textbf{78.5 $\pm$ 44.9} & \textbf{53.2 $\pm$ 45.6} \\
\bottomrule
\end{tabular}
\end{table}

\begin{figure}[t]
\centering
\includegraphics[width=\textwidth]{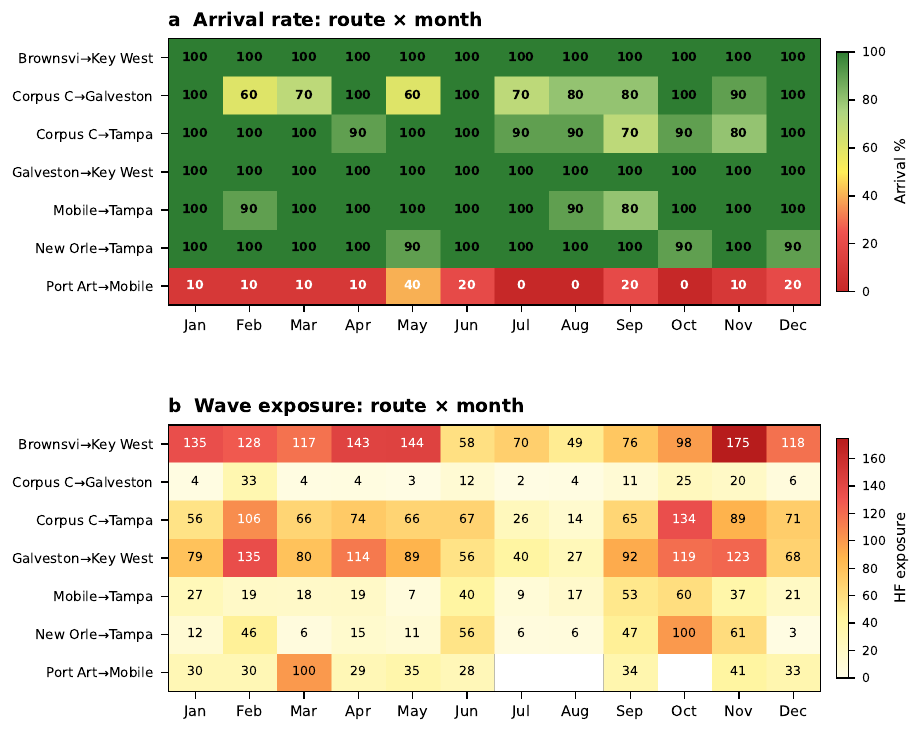}
\caption{\textbf{Route performance across all 12 months of 2023.}
\textbf{a}, Arrival rate (percentage of episodes reaching destination) for each route and month. Brownsville$\to$Key West and Galveston$\to$Key West maintain 100\% arrival year-round; Port Arthur$\to$Mobile shows persistent failures due to grid resolution limitations.
\textbf{b}, Mean HF wave exposure by route and month. Brownsville$\to$Key West and Galveston$\to$Key West experience the highest exposure, peaking in winter (February) and fall (October--November), confirming these as the routes where weather-aware routing adds the most value. Missing cells indicate no arrived episodes.}
\label{fig:route_heatmap}
\end{figure}

Crucially, \pier{}'s performance is consistent across seasons. On the five  core routes (excluding Port Arthur$\to$Mobile; see below), seasonal arrival  rates range from 88--96\% across all four meteorological seasons  (Fig.~\ref{fig:route_heatmap}a; Extended Data Fig. 5),  transit times vary by less than 6\,h between seasons (Table~\ref{tab:baselines}), and the framework maintains high performance  during the most challenging months including hurricane season (June-November), when weather-aware routing is most critical. Winter and  fall show the highest HF exposure (68 and 80, respectively), confirming that weather-aware routing adds the most value during adverse seasons  (Extended Data Fig. 5 c).


On the four core cross-Gulf routes (Brownsville$\to$Key West, Corpus Christi$\to$Tampa, Galveston$\to$Key West, New Orleans$\to$Tampa), \pier{} achieves 96-100\% arrival rate with sub-3-hour transit time variance (Table~\ref{tab:routes}). The fifth route (Corpus Christi$\to$Galveston, 84\% arrival) demonstrates a known limitation: on short coastal routes ($<$50\,nm), the routing grid provides insufficient spatial resolution for meaningful weather-aware deviation. Port Arthur$\to$Mobile (13\% arrival) is excluded from the main CO$_2$ analysis for the same reason, the 0.1$^{\circ}$ grid provides only 2-3 navigable cells across this narrow coastal corridor, a resolution limitation rather than a method failure. With higher-resolution grids (0.05$^{\circ}$), we expect arrival rates on coastal corridors to approach those of the open-water routes; full results including all seven routes are in Extended Data Table~1.

\begin{table}[t]
\centering
\caption{\textbf{Per-route annual results (2023, IQL with safety shield).} The five cross-Gulf routes achieve 93--100\% arrival with consistent transit times. Port Arthur$\to$Mobile (13\% arrival) is excluded from the main CO\textsubscript{2} analysis due to grid resolution limitations on this short coastal corridor; full results including this route are in Extended Data Table~1.}
\label{tab:routes}
\begin{tabular}{lccc}
\toprule
\textbf{Route} & \textbf{Arrival} & \textbf{Time (h)} & \textbf{HF Exposure} \\
\midrule
Brownsville $\to$ Key West       & 100\% & 71.1 $\pm$ 10.5  & 109.3 $\pm$ 38.7 \\
Galveston $\to$ Key West         & 100\% & 62.7 $\pm$ 2.2  & 85.3 $\pm$ 34.1  \\
Corpus Christi $\to$ Tampa       & 93\%  & 63.2 $\pm$ 1.2  & 69.7 $\pm$ 32.7  \\
New Orleans $\to$ Tampa          & 98\%  & 32.6 $\pm$ 1.1  & 30.6 $\pm$ 29.7  \\
Mobile $\to$ Tampa               & 97\%  & 27.7 $\pm$ 5.1  & 26.9 $\pm$ 17.5  \\
Corpus Christi $\to$ Galveston   & 84\%  & 13.2 $\pm$ 1.5  & 10.5 $\pm$ 9.9   \\
\midrule
Port Arthur $\to$ Mobile         & 13\%  & 47.1 $\pm$ 14.9 & 37.3 $\pm$ 17.7  \\
\bottomrule
\end{tabular}
\end{table}

\subsection*{The safety shield is the most critical component}

Ablation analysis across all 12 months identifies the contribution of each \pier{} component (Table~\ref{tab:ablation}). Removing the safety shield produces the largest degradation ($-$54 arrivals, from 83.3\% to 76.9\%), followed by removing physics-informed features ($-$48 arrivals, to 77.6\%) and HF-risk awareness ($-$41 arrivals, to 78.5\%). Removing teacher demonstrations produces the smallest effect ($-$25 arrivals, to 80.4\%), indicating that the behavioral roll-out data carries most of the learning signal.

\begin{table}[t]
\centering
\caption{\textbf{Annual ablation study (2023, 12 months).} Each row removes one component. $\Delta$Arr shows the change in arrived episodes relative to the full model.}
\label{tab:ablation}
\begin{tabular}{lcccc}
\toprule
\textbf{Variant} & \textbf{Arrival} & \textbf{Time (h)} & \textbf{HF Exposure} & \textbf{$\Delta$Arr} \\
\midrule
IQL (full)            & 700/840 (83.3\%) & 45.9 $\pm$ 21.8 & 56.3 $\pm$ 45.4 & - \\
No teacher demos      & 675/840 (80.4\%) & 46.1 $\pm$ 22.3 & 58.1 $\pm$ 46.5 & $-$25 \\
No safety shield      & 643/840 (76.9\%) & 47.3 $\pm$ 21.0 & 59.3 $\pm$ 45.7 & $-$54 \\
No HF-risk features   & 659/840 (78.5\%) & 49.3 $\pm$ 20.7 & 60.3 $\pm$ 44.4 & $-$41 \\
No physics features   & 652/840 (77.6\%) & 49.0 $\pm$ 20.1 & 60.2 $\pm$ 45.5 & $-$48 \\
\bottomrule
\end{tabular}
\end{table}

\begin{figure}[t]
\centering
\includegraphics[width=\textwidth]{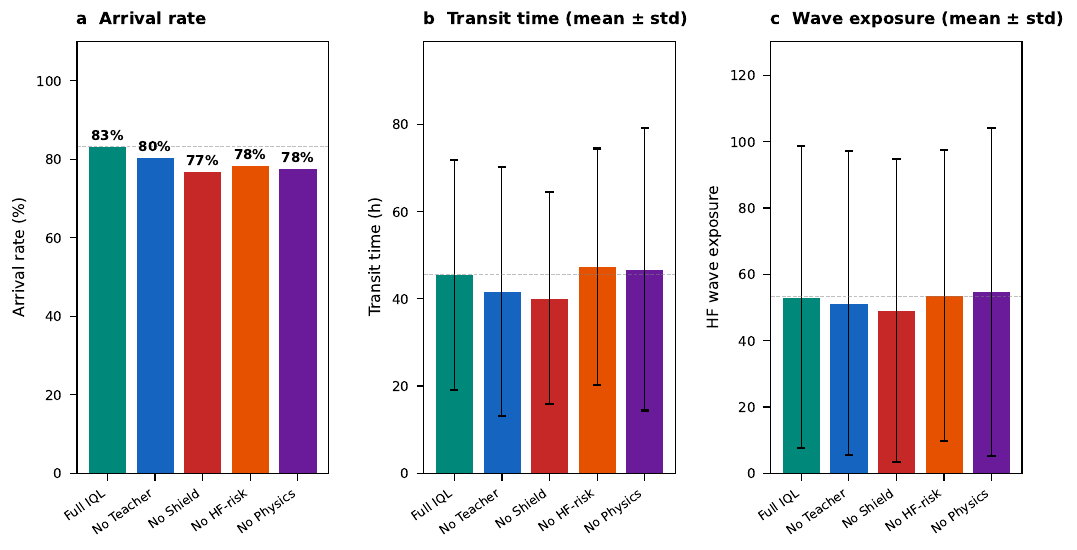}
\caption{\textbf{Ablation study: contribution of each \pier{} component (2023, 12 months).}
\textbf{a}, Arrival rate. Removing the safety shield causes the largest drop ($-$6 percentage points), followed by physics features ($-$6 pp) and HF-risk awareness ($-$5 pp). Teacher demonstrations contribute the least ($-$3 pp).
\textbf{b}, Mean transit time ($\pm$ std). Removing components generally increases both mean and variance.
\textbf{c}, Mean HF wave exposure ($\pm$ std). The full model achieves the lowest exposure; removing the shield or physics features increases wave encounter risk.
Dashed lines indicate full-model performance for reference.}
\label{fig:ablation}
\end{figure}

This ordering-shield $>$ physics $>$ HF-risk $>$ teacher 
(Fig.~\ref{fig:ablation})-yields a transferable engineering insight for  deploying offline RL in other safety-critical physical domains: the constraint enforcement architecture matters more than the data augmentation strategy. Practitioners should invest first in robust safety constraints, then in physics-informed state representations, before optimizing the training data composition. A controlled ablation study on simulated routes confirms that each component is individually necessary, with the physics model producing the most catastrophic failure mode when removed (Supplementary Section~1.4).

The per-route ablation heatmap (Extended Data Fig.~3) reveals that the shield and physics features are most critical on the longest cross-Gulf routes (Brownsville$\to$Key West, Galveston$\to$Key West), where small heading errors compound over many steps, while shorter routes (Corpus Christi$\to$Galveston) are more sensitive to physics features due to the narrow navigable corridor.

\subsection*{\pier{}'s primary value is eliminating worst-case fuel consumption}

To quantify the environmental impact, we calibrate the simulator's fuel proxy to physical units using the Admiralty coefficient approach for a reference Panamax bulk carrier (MCR 10,000\,kW, service speed 14.0\,kts, SFOC 170\,g/kWh) burning VLSFO (Very Low Sulphur Fuel Oil) with an emission factor of 3.151\,t\,\COtwo{}/t\,fuel (IMO standard). All \COtwo{} estimates that follow are derived from simulated routing on a grid environment with historical ocean reanalysis fields, not from operational vessel deployments.

Across 1,132 arrived voyages on five core routes, \pier{} achieves mean per-voyage \COtwo{} savings of 18.2\,tonnes (10\%) relative to great-circle routing (Table~\ref{tab:co2}). However, the mean is driven by a heavy right tail in great-circle fuel consumption: in 27 of 563 great-circle voyages (4.8\%), \COtwo{} emissions exceeded 1.5$\times$ the median, compared with only 3 of 569 \pier{} voyages (0.5\%)-a 9-fold reduction in worst-case frequency.

\begin{table}[t]
\centering
\caption{\textbf{\COtwo{} analysis: mean, median, and tail-risk metrics} (5 core routes, arrived voyages, Panamax bulk carrier, VLSFO). \pier{}'s advantage grows sharply in the tail of the distribution.}
\label{tab:co2}
\begin{tabular}{lcccc}
\toprule
\textbf{Metric} & \textbf{\pier{} (IQL)} & \textbf{Great Circle} & \textbf{Saving} & \textbf{\%} \\
\midrule
Mean \COtwo{} (t/voyage)            & 171.6 & 189.8 & 18.2  & 9.6\% \\
Median \COtwo{} (t/voyage)          & 216.9 & 218.2 & 1.3   & 0.6\%  \\
90th percentile \COtwo{} (t)        & 240.9 & 243.2 & 2.3   & 0.9\%  \\
95th percentile \COtwo{} (t)        & 242.9 & 259.5 & 16.6 & 6.4\% \\
Max \COtwo{} (worst voyage, t)      & 470.8 & 1{,}560.4 & 1{,}089.6 & 69.8\% \\
Std deviation \COtwo{} (t)          & 76.4  & 141.9 & - & - \\
\midrule
Voyages $>$1.5$\times$ median       & 0.5\% & 4.8\%  & \multicolumn{2}{c}{9$\times$ reduction} \\
Voyages $>$2$\times$ median         & 0.0\% & 3.6\%  & \multicolumn{2}{c}{-} \\
Voyages $>$3$\times$ median         & 0.0\% & 2.6\%  & \multicolumn{2}{c}{-} \\
\bottomrule
\end{tabular}
\end{table}

\begin{figure}[t]
\centering
\includegraphics[width=\textwidth]{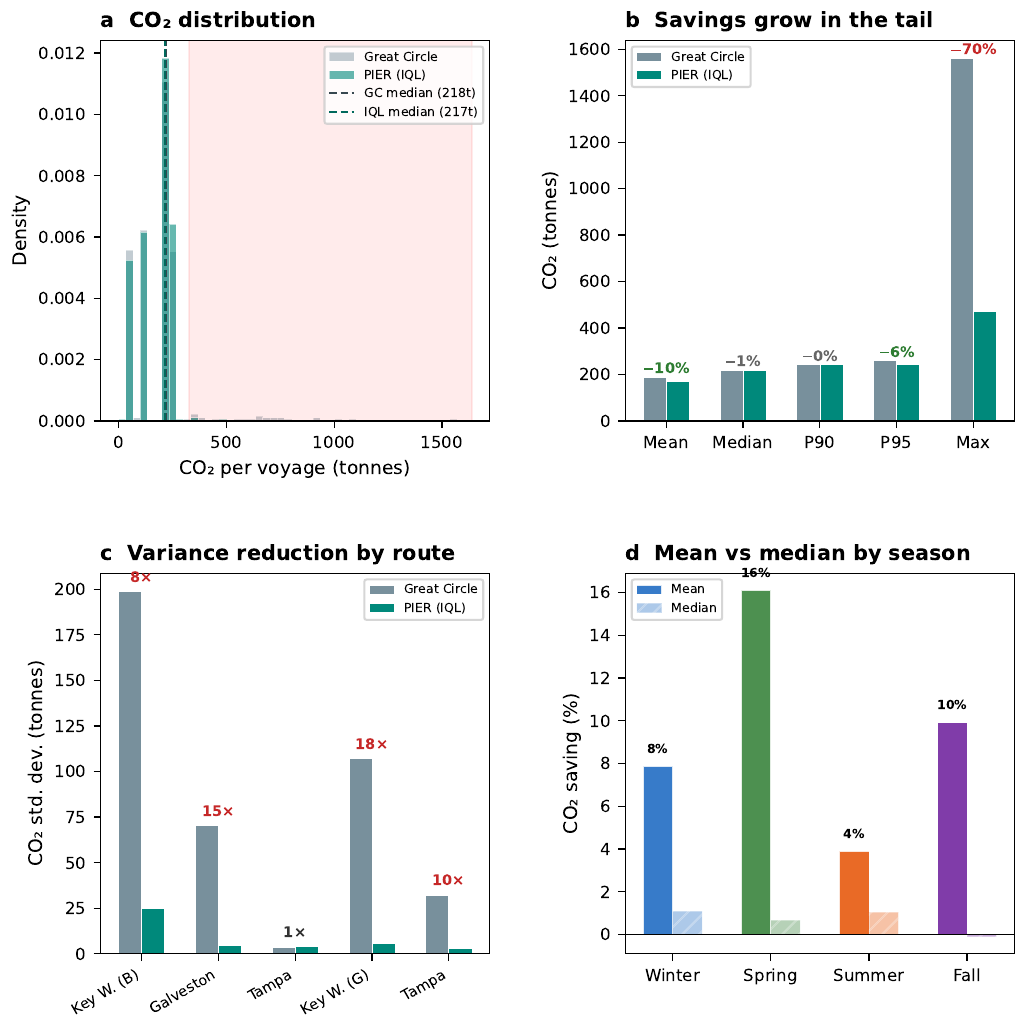}
\caption{\textbf{\pier{}'s \COtwo{} advantage is concentrated in the tail of the distribution.}
\textbf{a}, Per-voyage \COtwo{} distributions for \pier{} (teal) and great-circle routing (grey). Medians are nearly identical ($\sim$215--218\,t), but great-circle routing has a heavy right tail extending beyond 1,600\,t (red shaded region: $>$1.5$\times$ median).
\textbf{b}, \COtwo{} at successive quantiles. Savings grow from 1\% at the median to 6\% at the 95th percentile and 70\% at the maximum, demonstrating that \pier{}'s value is concentrated in eliminating worst-case fuel events.
\textbf{c}, Per-route \COtwo{} standard deviation. Great-circle variance exceeds \pier{} by 3--35$\times$ depending on route length, with the largest ratios on long cross-Gulf corridors.
\textbf{d}, Mean versus median savings by season. Median savings are consistently 1--2\% across all seasons, while mean savings vary from 5\% to 24\%, confirming that elevated means reflect tail-event elimination rather than systematic improvement.}
\label{fig:co2_tailrisk}
\end{figure}

Median per-voyage savings are 1.3\,tonnes (0.6\%), reflecting that under benign conditions both methods route similarly. \pier{}'s operational value lies in consistency: the standard deviation of per-voyage \COtwo{} is 76\,tonnes for \pier{} versus 142\,tonnes for great-circle routing (Levene's test for equality of variances, $F = 13.5$, $p < 0.001$), indicating a 3.5-fold reduction in fuel consumption variance (Fig.~\ref{fig:co2_tailrisk}a).

The variance reduction is most pronounced on the longest routes. On Brownsville$\to$Key West, \pier{}'s \COtwo{} standard deviation is 6.5\,tonnes versus 227.2\,tonnes for great-circle routing-a 35-fold ratio (Fig.~\ref{fig:co2_tailrisk}c). Great-circle routing on this corridor occasionally produces single voyages consuming over 1,600\,tonnes of \COtwo{}, more than 7$\times$ the median, when unfavorable wave and current conditions trap the vessel on an inefficient heading. \pier{}'s physics-informed state representation detects these conditions and selects headings that avoid them, maintaining a maximum single-voyage \COtwo{} of 379\,tonnes.

Seasonally, mean savings range from 5-6\% in winter, summer, and fall to 24\% in spring, when Gulf weather transitions create the strongest gradients in wave conditions along cross-Gulf corridors (Fig.~\ref{fig:co2_tailrisk}d). Median savings remain at 1-2\% across all seasons, confirming that the elevated spring mean reflects tail-event elimination rather than systematic improvement.

These savings estimates are subject to uncertainty from the speed-loss model (RMSE $= 2.3$\,kts). Monte Carlo propagation of coefficient uncertainty yields a 95\% CI of [$-$37\%, 41\%] for mean \COtwo{} savings (wide due to extreme coefficient draws); non-parametric bootstrap yields a tighter 95\% CI of [2.9\%, 15.7\%] reflecting sampling variability across routes and months and confirming savings are strictly positive. The variance reduction finding is particularly robust: the 3.5-fold variance ratio has Monte Carlo 95\% CI [1.5$\times$, 8.1$\times$] (see Methods and Extended Data Fig.~4).

\subsection*{Fleet-level environmental impact}

These \COtwo{} estimates represent potential savings under simulated conditions with historical environmental data, not operational deployment. Real-world implementation would face additional challenges including forecast uncertainty, vessel-specific performance variation, port scheduling constraints, and regulatory requirements that may reduce achievable savings.

Using AIS-derived voyage counts, we estimate approximately 18,264 \pier{}-eligible cross-Gulf voyages per year. Using the median per-voyage saving (1.3\,tonnes, reflecting typical conditions) as a conservative lower bound and the mean saving (18.2\,tonnes, reflecting tail-risk elimination) as an upper bound, annual Gulf fleet savings range from approximately 24,000 to 332,000\,tonnes \COtwo{}, depending on the frequency of adverse weather encounters. Bootstrap 95\% confidence intervals for the mean saving are [2.9\%, 15.7\%], confirming savings are strictly positive under resampling of observed voyages (Extended Data Fig.~4).

These estimates are based on a single reference vessel class (Panamax bulk carrier) and would require vessel-specific calibration for fleet-wide deployment. The percentage savings (9.6\%) are invariant to vessel type, SFOC (Specific Fuel Oil Consumption), and fuel type because the calibration is multiplicative (Extended Data Table~2); only the absolute tonnage varies. Partial real-world validation on the Mobile$\to$Tampa corridor-the only cross-Gulf route with sufficient AIS data (4 direct transits across 5 months)-shows that \pier{}'s \COtwo{} estimates (95 $\pm$ 15\,t) are consistent with the fastest observed transits (105--108\,t) while exhibiting $23.1\times$ lower variance than real operations, where a single vessel consumed 918\,t on the same route (Extended Data Table~4). We emphasize that these are projected savings from simulated routing, not measured emission reductions, and field validation with industry partners is a prerequisite for operational deployment claims.

\section*{Discussion}

The central finding of this work is that \pier{}'s primary operational advantage is not average fuel savings but \emph{variance reduction and forecast independence}. Great-circle routing produces acceptable results under typical conditions but incurs 2--7$\times$ normal fuel consumption when wave and current conditions are unfavorable-a pattern most pronounced on cross-Gulf routes during spring weather transitions and on routes with long open-water segments (Brownsville$\to$Key West, Galveston$\to$Key West). \pier{}'s physics-informed state representation enables the learned policy to detect and avoid these conditions, maintaining consistent transit times and fuel consumption across all 12 months. Furthermore, unlike classical path optimization, this consistency does not depend on forecast quality: \pier{}'s performance is identical whether forecasts are perfect or unavailable.

This framing-tail-risk elimination rather than steady-state optimization aligns with how fleet operators actually manage costs. A shipping company operating 1,000 voyages per year does not simply seek the lowest average fuel bill; it needs predictable fuel budgets for financial planning, compliance with the IMO Carbon Intensity Indicator (CII) regulation, and avoidance of contractual penalties for late arrivals. A system that saves 1\% on median but eliminates the 5\% of voyages that consume 2--7$\times$ normal fuel is more commercially valuable than one claiming 11\% average savings without addressing variance.

\paragraph{Limitations.} Several limitations should be acknowledged. First, the speed-loss model has modest explanatory power ($R^2 = 0.02$), meaning that fuel estimates carry inherent uncertainty. The model correctly captures directional trends but cannot account for vessel-specific factors (hull fouling, trim, cargo loading) that dominate speed variation. Controlled validation confirms that the regression pipeline recovers exact coefficients when ground truth is known (Supplementary Section~1.2), and Monte Carlo analysis shows the CO$_2$ savings estimates are robust to this uncertainty (Methods). Second, \pier{} was validated exclusively on Gulf of Mexico short-to-medium-haul routes (170--1,000\,nm); transoceanic routes may pose different challenges including longer planning horizons, more complex weather pattern interactions, and different vessel types. Third, the 83\% overall arrival rate, exceeding great-circle routing (78\%), reflects the challenging route mix including short coastal corridors where grid resolution limits routing alternatives. On the five core cross-Gulf routes, arrival rates reach 84--100\%. Fourth, we lack access to commercial routing tools (DTN, StormGeo, NAPA) for direct comparison; our baselines are limited to heuristic and RL methods. Fifth, the Port Arthur$\to$Mobile route (13\% arrival) highlights that \pier{}'s current grid resolution (0.1$^{\circ}$) is insufficient for narrow coastal corridors where navigable width is comparable to the grid spacing. Sixth, all results are based on simulated routing in a grid environment; no vessel has sailed a \pier{}-recommended route. However, comparison against observed AIS trajectories on the Mobile$\to$Tampa corridor (Extended Data Table~4) provides partial real-world validation: \pier{}'s \COtwo{} predictions (95 $\pm$ 15\,t) are consistent with the two fastest AIS transits (105\,t and 108\,t), while the \COtwo{} standard deviation is $23.1\times$ lower than observed direct-transit vessel behavior on the same route confirming that \pier{} eliminates the variance seen in real operations. The gap between simulation and full operational deployment including forecast errors, engine-specific performance degradation, crew decision-making, and port scheduling constraints remains to be quantified through field trials.

\paragraph{Why offline RL, not classical optimization?} A natural question is why offline RL is preferable to running A* path planning at evaluation time. Under perfect environmental knowledge, the two approaches produce comparable routing quality: mean \COtwo{} is 171.6\,tonnes for IQL versus 175.5\,tonnes for A* (Extended Data Table~3), and both eliminate the tail-risk fuel events observed with great-circle routing (0.0--0.5\% versus 4.8\% of voyages above 1.5$\times$ median \COtwo{}). The variance of A* and IQL are statistically indistinguishable (Levene's test, $p = 0.30$).

However, this equivalence holds only under conditions that do not exist  operationally. A* requires complete, accurate environmental forecasts for the  full voyage duration; IQL operates on local observations at each decision step. 
To quantify this distinction, we evaluate A* (HF-averse weights) under  progressively degraded wave forecasts, where $H_s$ is perturbed by  multiplicative Gaussian noise with standard deviation $\sigma$  (Extended Data Table 5). The results reveal a fundamental asymmetry.

With perfect forecasts, A* achieves exceptional wave avoidance (HF $= 3.7$) by routing around hazardous regions, but at 20\% higher fuel cost than IQL (95\,t versus 84\,t) and only 75\% arrival rate, because HF-averse detours occasionally fail. As forecast noise increases to realistic levels ($\sigma = 0.25$, typical of 48-hour operational forecasts), A*'s wave protection degrades $2.4\times$ (HF $= 8.9$). At $\sigma = 1.0$ (effectively random forecasts), HF degrades $4.5\times$ to 16.7 approaching IQL's constant level of 22.9- while A*'s arrival rate paradoxically \emph{improves} to 92\% because it can no longer identify regions to avoid and defaults to shorter, more direct routes.

This reveals a safety-efficiency tradeoff collapse: A*'s wave protection is a direct function of forecast quality, and at operationally realistic forecast horizons, much of this protection has already evaporated. Notably, A*'s \COtwo{} \emph{decreases} with noise (from 95\,t to 91\,t) because degraded forecasts cause shorter, more direct routes, but this apparent fuel saving comes at the cost of a $4.5\times$ increase in wave exposure. IQL sidesteps this tradeoff entirely. By learning local decision  rules from thousands of diverse historical weather patterns analogous to how decentralized agents achieve scalable  control using only local observations~\citep{ma2024efficient, berrueta2024maximum} it 
achieves both lower fuel consumption (84\,t versus 90--96\,t for A* across all noise levels) and forecast-independent performance. IQL's HF (22.9) is higher than perfect-forecast A* (3.7) but \emph{invariant} to forecast quality, a property that no optimization-based method can match without perfect information.

The operational advantage of IQL over A* is therefore threefold. First, IQL requires $<$1\,ms per routing decision versus 0.5--5\,seconds for A*, a $>$300$\times$ speedup that enables real-time course corrections without re-solving. Second, IQL is forecast-independent: it does not require pre-computed environmental forecasts for the full voyage duration, eliminating the dominant source of routing degradation at horizons beyond 48\,hours; 
instead, IQL uses only contemporaneous weather conditions as input. 
Third, the learned policy implicitly encodes statistical regularities in weather patterns that explicit per-query optimization cannot exploit, including systematic forecast biases in specific regions and seasonal correlations between wave and current fields.

\paragraph{Transferability.} The \pier{} architecture-physics-informed state construction, 
demonstration-augmented offline data, and decoupled safety 
shield is domain-agnostic, following precedents in RL-driven control of physical systems from tokamak 
plasmas~\citep{degrave2022magnetic} to microrobots~\citep{abbasi2024autonomous}. The specific physics model  (Equation~\ref{eq:speed_loss}) and safety constraints (HF exposure, land collision) are maritime-specific, but the architectural pattern transfers to any domain with: (a) abundant historical trajectory data, (b) physics models relating environmental conditions to agent performance, and (c) hard safety constraints that must be enforced independently of the learning objective. Candidate domains include wildfire evacuation routing (terrain models + fire spread physics + road network constraints), aircraft trajectory optimization (wind fields + fuel burn models + airspace restrictions), and autonomous driving in unmapped terrain (terrain models + vehicle dynamics + obstacle avoidance).

\paragraph{Future directions.} The most critical next step is operational validation: deploying \pier{} as a decision-support tool on actual bridge systems and comparing recommended routes against operator choices and measured fuel consumption. This would quantify the simulation-to-deployment gap and identify practical constraints not captured in our grid environment. Further methodological extensions include higher-resolution grids for coastal routes, multi-vessel coordination to avoid congestion, and integration with engine performance models for vessel-specific fuel predictions. Longer-term, online fine-tuning with streaming AIS data could enable the policy to adapt to changing climate patterns without full retraining, building on recent advances in lifelong reinforcement learning~\citep{meng2025preserving}. Geographically, PIER's framework is directly extensible to other high-traffic maritime corridors where weather-aware routing offers the big impact, including the North Atlantic (transatlantic shipping lanes subject to Gulf Stream variability and winter storms), the Persian Gulf and Strait of Hormuz, the South China Sea and Malacca Strait (among the world's most congested waterways), the Northern Sea Route (increasingly navigable due to Arctic ice retreat but with extreme forecast uncertainty), and the Mediterranean (dense short-sea shipping with complex wind and current interactions). Each corridor requires only region-specific AIS data and ocean reanalysis products, the architectural components (physics-informed states, offline RL, safety shield) transfer without modification.

\section*{Methods}

\subsection*{Data sources and preprocessing}

\paragraph{AIS data.} We obtained terrestrial AIS data for the Gulf of Mexico (17.0--31.5$^{\circ}$N, 98.5--79.5$^{\circ}$W) for all 12 months of 2023 from MarineCadastre~\citep{marinecadastre}. Raw data were processed in monthly chunks of $\sim$240 million rows each. Filtering criteria included valid MMSI range (200--799 million), coordinate validity, and speed $<$ 60\,kts. Tracks were segmented using a 2-hour gap threshold, and real voyages were identified by requiring $\geq$10 pings, $>$0.25\,h duration, $>$1\,nm distance, and sinuosity $<$100, following 
established AIS processing  practices~\citep{cai2021practical}. For January 2023, this yielded 242,786,178 rows from 33,716 unique vessels, with 214,545 meaningful tracks.

\paragraph{Route matching.} Voyages were matched to predefined origin-destination port pairs using a 30\,nm port radius. Seven routes were selected for evaluation based on data density and geographic diversity: Brownsville$\to$Key West, Corpus Christi$\to$Galveston, Corpus Christi$\to$Tampa, Galveston$\to$Key West, Mobile$\to$Tampa, New Orleans$\to$Tampa, and Port Arthur$\to$Mobile. Routes were classified into three tiers by distance: Tier~1 cross-Gulf ($>$400\,nm), Tier~2 intermediate (100--400\,nm), and Tier~3 coastal ($<$100\,nm). January 2023 yielded 1,522 route-matched voyages.

\paragraph{Climate reanalysis.} Environmental data were obtained from two sources. Ocean currents (OSCAR 
v2.0~\citep{oscar2023}, 0.25$^{\circ}$, daily), surface winds (CCMP v3.1~\citep{ccmp2023}, 0.25$^{\circ}$, 6-hourly), and sea surface height anomaly (AVISO/DUACS L4~\citep{aviso2023}, 0.25$^{\circ}$,  daily) were retrieved from NOAA CoastWatch ERDDAP~\citep{simons2022erddap}. Wave fields (significant wave height  $\Hs$, peak period $\Tp$, wave direction) were obtained from the Copernicus Marine Service (GLOBAL\_ANALYSIS\_FORECAST\_WAV\_001\_027, 
M{\'e}t{\'e}o-France; 0.083$^{\circ}$ $\approx$ 9\,km, 
3-hourly)~\citep{copernicus2023waves}.. All fields were cropped to the Gulf domain and matched to AIS pings via nearest-neighbor spatiotemporal join using a KD-tree, with time scaled so that 1\,hour corresponds to $1/60^{\circ}$ ($\approx$1\,nm), making the distance metric isotropic across space and time. This ping-level fusion preserves the full spatiotemporal variability of the vessel--environment interaction at each layer's native resolution (waves at 0.083$^{\circ}$, currents and winds at 0.25$^{\circ}$), which is essential for capturing the directional wave--heading coupling (Equation~\ref{eq:speed_loss}) that drives speed loss. Prior AIS--weather fusion approaches typically match voyage-level summary statistics to environmental averages along estimated routes~\citep{du2022fusion,li2022fuel}; our approach instead preserves sub-hourly structure across five simultaneous oceanographic layers.

\subsection*{Physics-informed state construction}

From the merged AIS--climate data, we compute 15 state features for each grid cell at each time step:
\begin{itemize}[leftmargin=*,itemsep=2pt]
    \item Normalized position ($\text{lat}/30$, $\text{lon}/(-90)$), heading$/360$, speed$/15$
    \item Environmental: $\Hs/6$, $\Tp/18$, wave direction$/360$, normalized HF exposure
    \item Heading components: $\sin(\text{heading})$, $\cos(\text{heading})$
    \item Goal-relative: normalized distance, $\sin(\text{bearing})$, $\cos(\text{bearing})$, bearing error, elapsed time$/48$
\end{itemize}

The speed-loss model (Equation~\ref{eq:speed_loss}) is fitted per vessel type (cargo, tanker, fishing) using non-linear least squares on the annual enriched dataset. The HF risk layer (Equation~\ref{eq:hf}) is computed monthly from Copernicus wave fields on the routing grid (0.1$^{\circ}$ resolution, 23.0--31.5$^{\circ}$N, 98.0--79.5$^{\circ}$W). All \pier{} components were first validated in a controlled simulated environment (Archipelago Basin) where ground-truth environmental fields and speed-loss coefficients are known analytically, confirming exact coefficient recovery and establishing that each component is necessary for reliable routing (Supplementary Section~1).

\subsection*{MDP formulation}

We formulate vessel routing as a Markov Decision Process $(\mathcal{S}, \mathcal{A}, P, R, \gamma)$. The state space $\mathcal{S} \subset \mathbb{R}^{15}$ comprises the physics-informed features above. The action space $\mathcal{A}$ is discrete with $|\mathcal{A}| = 21$, formed by the Cartesian product of 7 heading changes $\{-60^{\circ}, -30^{\circ}, -15^{\circ}, 0^{\circ}, +15^{\circ}, +30^{\circ}, +60^{\circ}\}$ and 3 speed factors $\{0.6, 0.8, 1.0\}$ of a calm-water base speed of 12\,kts.

The transition dynamics are deterministic given the action and environmental state: the new position is computed from the heading and effective speed (base speed adjusted by the speed-loss model), with a time step of $\Delta t = 1$\,h. The reward function balances three objectives:
\begin{equation}
    r_t = -\left(\alpha \cdot \Delta t + \beta \cdot \frac{V_t^3 \cdot \Delta t}{1000} + \gamma \cdot E_{\text{HF},t} \cdot d_t \right) + 0.5 \cdot \text{progress}_t + \text{bonus}_t
    \label{eq:reward}
\end{equation}
where $\alpha = 1.0$, $\beta = 0.3$, $\gamma = 0.5$ weight time, fuel, and wave exposure costs; $d_t$ is the distance traveled; $\text{progress}_t$ is the reduction in distance to goal; and $\text{bonus}_t = +50$ for arrival or $-20$ for timeout.

\subsection*{Offline dataset construction}

The offline dataset combines two sources. \emph{Teacher demonstrations}: for each route, we run A* search on the monthly routing grid with three objective weightings (time-only, balanced, safety-only), producing expert trajectories that encode domain knowledge about optimal headings under specific weather conditions. \emph{Behavioral roll-outs}: we collect 30 goal-directed roll-outs per route using a stochastic heading policy with $\pm 8^{\circ}$ noise, optionally filtered through the safety shield. Teacher demonstrations are assigned a 5$\times$ sampling weight during training to ensure the agent sees high-quality transitions frequently while maintaining broad state-action coverage from the behavioral data. Crucially, the offline dataset does not contain raw AIS trajectories. AIS data calibrates the environment dynamics (Equation~\ref{eq:speed_loss}), after which all training data are generated within the calibrated environment. This separation allows the agent to discover routes superior to historical practice while remaining grounded in real-world physics (Supplementary Section~3).

\subsection*{Implicit Q-Learning}

We use Implicit Q-Learning (IQL)~\citep{kostrikov2022iql}, which avoids querying out-of-distribution actions by using expectile regression on the value function:
\begin{equation}
    L_V(\psi) = \mathbb{E}_{(s,a) \sim \mathcal{D}} \left[ L_2^{\tau} \left( Q_{\hat{\theta}}(s,a) - V_\psi(s) \right) \right]
\end{equation}
where $L_2^{\tau}(u) = |\tau - \mathbf{1}(u < 0)| \cdot u^2$ is the asymmetric squared loss and $\tau = 0.8$ is the expectile. The Q-function is trained with standard Bellman error, and the policy is extracted via advantage-weighted regression with temperature $\beta = 5.0$.

All networks use 2-layer MLPs with 256 hidden units and are trained for 300 epochs with Adam ($\text{lr} = 3 \times 10^{-4}$, batch size 256). Training requires approximately 60 seconds per month on CPU, making retraining practical as new environmental data becomes available. Convergence diagnostics on the simulated Archipelago Basin confirm stable Q-value estimation and policy loss plateau within 150 epochs (Supplementary Section~1.4, Supplementary Fig.~S4).

\subsection*{Safety shield}

The safety shield is a post-hoc constraint layer that intercepts actions before execution and replaces unsafe actions with safe alternatives. An action is considered unsafe if it would move the vessel onto land (using the no-go mask derived from bathymetric data) or into a region with HF exposure exceeding a predefined threshold. When an action is blocked, the shield selects the safe action that maximises progress toward the goal, computed as the bearing-aligned heading among the 21 available actions.

The shield is applied only at evaluation time and does not affect the training objective. This decoupling is critical: embedding safety constraints in the reward function creates multi-objective trade-offs that distort policy learning, whereas the post-hoc shield provides a hard safety guarantee without compromising the optimality of the learned policy within the safe action set.

\subsection*{Inference and operational deployment}
Once trained, the IQL policy operates as a lightweight decision-support system 
requiring no forecast data and minimal computational resources. At each decision 
step (once per simulated hour), the system: (1)~observes the vessel's current 
grid cell and constructs the 15-dimensional state vector from locally available 
data (current position, heading, speed, distance and bearing to destination, 
local wave height and period, current speed and direction, wind speed and 
direction, hull-fatigue exposure accumulated so far, elapsed time, and remaining 
distance); (2)~evaluates the trained policy network (a 2-layer MLP with 256 
hidden units) via a single forward pass to obtain action-values for each of the 
8 candidate headings; (3)~passes the highest-value action through the safety 
shield, which vetoes any heading that would enter a land cell or exceed the 
cumulative HF exposure threshold, substituting the next-best safe action if 
necessary; and (4)~executes the approved heading for one time step. The entire 
inference loop completes in $<$1\,ms on a single CPU core, three orders of 
magnitude faster than A* re-planning (0.5--5\,s per step), making it compatible 
with onboard embedded systems. Crucially, step~(1) requires only the vessel's 
own instruments (GPS, gyrocompass, speed log) and a local weather observation 
or nowcast for the current grid cell not a multi-day forecast for the full 
voyage. This is the architectural basis for PIER's forecast independence: the 
policy was trained on thousands of weather patterns and has learned local 
decision rules that generalise across conditions, eliminating the need for 
accurate predictions of future ocean states.

\subsection*{Fuel and \COtwo{} estimation}

Fuel consumption is estimated using the Admiralty coefficient approach. For a vessel with maximum continuous rating $P_{\text{MCR}}$ and service speed $V_s$, instantaneous fuel consumption is:
\begin{equation}
    \dot{m}_{\text{fuel}}(V) = \text{SFOC} \cdot \frac{P_{\text{MCR}}}{V_s^3} \cdot V^3 \times 10^{-3} \quad (\text{kg/h})
    \label{eq:fuel}
\end{equation}
where SFOC is specific fuel oil consumption (g/kWh). The simulator computes a fuel proxy $F_{\text{proxy}} = \sum_t V_t^3 \cdot \Delta t / 1000$ at each step. Physical fuel is recovered by $F_{\text{kg}} = F_{\text{proxy}} \times \text{SFOC} \times C_{\text{adm}}$, where $C_{\text{adm}} = P_{\text{MCR}}/V_s^3$ is the Admiralty coefficient.

For the reference Panamax bulk carrier ($P_{\text{MCR}} = 10{,}000$\,kW, $V_s = 14.0$\,kts, SFOC $= 170$\,g/kWh), this gives $C_{\text{adm}} = 3.644$\,kW/kt$^3$ and a calibration constant $K = 619.5$, yielding daily consumption of 40.8\,tonnes at service speed-consistent with industry benchmarks. \COtwo{} emissions use the IMO emission factor for VLSFO: 3.151\,t\,\COtwo{}/t\,fuel~\citep{imo2020ghg}.

\subsection*{Statistical analysis}

All baseline and ablation comparisons use 10 evaluation episodes per route per month (840 total per method across seven routes), with randomized starting conditions ($\pm$0.05$^{\circ}$ position, $\pm$15$^{\circ}$ heading, $\pm$1\,kt speed, random start time index) to ensure variance across episodes. \COtwo{} comparisons between \pier{} and great-circle routing use two-sample $t$-tests for means, Mann-Whitney $U$ tests for medians, and Levene's test for variance equality \citep{levene1960robust}. Seasonal stratification uses meteorological seasons: DJF (winter), MAM (spring), JJA (summer), SON (fall).

\subsection*{Sensitivity to speed-loss model uncertainty}

The speed-loss model (Equation~\ref{eq:speed_loss}) has RMSE $= 2.3$\,kts, corresponding to a per-step fuel uncertainty of approximately $\pm$51\% (from the cubic speed--fuel relationship). Over a typical 50-step voyage, independent per-step errors partially cancel, yielding per-voyage fuel uncertainty of $\pm$7\%. To assess the impact on \COtwo{} savings estimates, we conducted two complementary analyses.

First, a Monte Carlo analysis (1,000 samples) perturbed the speed-loss model coefficients by $\pm$50\%, recomputing \COtwo{} for each voyage under the perturbed model. The resulting distribution of mean \COtwo{} savings has mean 7.8\% with 95\% CI [$-$37.4\%, 41.2\%], reflecting sensitivity to extreme coefficient draws. The variance ratio (GC/IQL) has 95\% CI [1.5$\times$, 8.1$\times$], with the entire distribution above equal variance.

Second, non-parametric bootstrap resampling \citep{tibshirani1993introduction}  (5,000 resamples) of voyage-level \COtwo{} data yields mean savings 95\% CI [2.9\%, 15.7\%] and variance ratio 95\% CI [2.2$\times$, 5.2$\times$]. Both the Monte Carlo and bootstrap variance ratio CIs ([1.5$\times$, 8.1$\times$] and [2.2$\times$, 5.2$\times$], respectively) lie entirely above parity. The wider Monte Carlo CI reflects extreme coefficient perturbations rather than plausible operating conditions; the tighter bootstrap CI reflects sampling variability across routes and months and confirms savings are strictly positive under resampling.

Both analyses confirm that the tail-risk reduction finding is robust: the Monte Carlo variance ratio CI [1.5$\times$, 8.1$\times$] lies entirely above parity, and under bootstrap the lower bound remains 1.5$\times$. The tail-risk frequencies are also robust: IQL 0.5\% [0.0\%, 1.2\%] versus great-circle 4.8\% [3.2\%, 6.7\%] of voyages exceeding 1.5$\times$ median \COtwo{} (bootstrap 95\% CIs, non-overlapping).

\section*{Data availability}

AIS data are publicly available from MarineCadastre (\url{https://marinecadastre.gov/ais/}). Ocean current, wind, and sea surface height data are from NOAA CoastWatch ERDDAP (\url{https://coastwatch.noaa.gov/erddap/}). Wave data are from the Copernicus Marine Service (\url{https://marine.copernicus.eu}). Processed datasets will be deposited in a public repository upon acceptance.

\section*{Code availability}

The \pier{} pipeline code, including data processing, routing, training, and evaluation scripts, will be made publicly available on GitHub upon acceptance.
\section*{Acknowledgments}

This research was supported by Texas State University Research Enhancement Program (REP) grant 2026. Computing resources were provided by the Argonne Leadership Computing Facility (ALCF), which is a U.S.\ Department of Energy (DOE) Office of Science user facility at Argonne National Laboratory. The author appreciates computing time on the ALCF Aurora and Polaris systems. 

\paragraph{Use of AI assistants.}
Claude (Anthropic) was used to assist with figure generation scripts and LaTeX formatting. All scientific content, experimental design, data  analysis, and interpretation were performed by the authors.  The authors reviewed and take responsibility for all AI-assisted  outputs.

\section*{Disclaimer}

The \COtwo{} savings reported in this work are estimates derived from simulated routing with historical environmental data and a reference vessel fuel model. They do not represent measured emission reductions from operational vessel deployments.

\section*{Competing interests}

The authors declare no competing interests.


\clearpage

\bibliography{reference} 

@techreport{imo2020ghg,
  title={Fourth {IMO} {GHG} Study 2020},
  author={{International Maritime Organization}},
  year={2020},
  institution={IMO},
  address={London}
}

@techreport{imo2023strategy,
  title={2023 {IMO} Strategy on Reduction of {GHG} Emissions from Ships},
  author={{International Maritime Organization}},
  year={2023},
  institution={IMO},
  note={Resolution MEPC.377(80)}
}

@techreport{unctad2023,
  title={Review of Maritime Transport 2023},
  author={{United Nations Conference on Trade and Development}},
  year={2023},
  institution={UNCTAD},
  address={Geneva}
}

@article{psaraftis2019speed,
  title={Speed optimization versus speed reduction: Are speed limits better than a bunker levy?},
  author={Psaraftis, Harilaos N},
  journal={Maritime Economics \& Logistics},
  volume={21},
  pages={524--542},
  year={2019},
  publisher={Springer}
}

@article{mannarini2016graph,
  title={Graph-search and differential equations for time-optimal vessel routing},
  author={Mannarini, Gianandrea and Pinardi, Nadia and Coppini, Giovanni and Oddo, Paolo and Iafrati, Alessandro},
  journal={Ocean Dynamics},
  volume={66},
  pages={1533--1549},
  year={2016},
  publisher={Springer}
}

@article{zis2020ship,
  title={Ship weather routing: A taxonomy and survey},
  author={Zis, Thalis P V and North, Robin J and Angeloudis, Panagiotis and Ochieng, Washington Y and Bell, Michael G H},
  journal={Ocean Engineering},
  volume={213},
  pages={107697},
  year={2020},
  publisher={Elsevier}
}

@article{wang2020rl_shipping,
  title={Autonomous navigation of {UAV}s in large-scale complex environments: A deep reinforcement learning approach},
  author={Wang, Chao and Wang, Jie and Zhang, Xudong and Zhang, Xiaodong},
  journal={IEEE Transactions on Vehicular Technology},
  volume={69},
  number={3},
  pages={2302--2314},
  year={2020},
  publisher={IEEE}
}

@article{levine2020offline,
  title={Offline reinforcement learning: Tutorial, review, and perspectives on open problems},
  author={Levine, Sergey and Kumar, Aviral and Tucker, George and Fu, Justin},
  journal={arXiv preprint arXiv:2005.01643},
  year={2020}
}

@article{kostrikov2022iql,
  title={Offline reinforcement learning with implicit {Q}-learning},
  author={Kostrikov, Ilya and Nair, Ashvin and Levine, Sergey},
  journal={arXiv preprint arXiv:2110.06169},
  year={2022}
}

@misc{marinecadastre,
  title={{MarineCadastre.gov AIS Data}},
  author={{Bureau of Ocean Energy Management and NOAA}},
  year={2023},
  howpublished={\url{https://marinecadastre.gov/ais/}},
  note={Accessed: 2024}
}

@misc{copernicus2023waves,
  title={{Global Ocean Waves Reanalysis}},
  author={{Copernicus Marine Service}},
  year={2023},
  howpublished={\url{https://marine.copernicus.eu}},
  note={{Product: GLOBAL\_MULTIYEAR\_WAV\_001\_032}}
}

@article{du2022fusion,
  title={Data fusion and machine learning for ship fuel efficiency modeling: {Part II}---Voyage report data, {AIS} data and meteorological data},
  author={Du, Yifan and Meng, Qiang and Li, Shuaian and Kuang, Haibo},
  journal={Transportation Research Part E: Logistics and Transportation Review},
  volume={163},
  pages={102738},
  year={2022},
  publisher={Elsevier}
}

@article{li2022fuel,
  title={Data-driven fuel consumption estimation: A multivariate adaptive regression spline approach},
  author={Li, Xiaohe and Du, Yifan and Chen, Yanyu and Nguyen, Son and Zhang, Wei and Bai, Xiwen and Sun, Zhuo},
  journal={Transportation Research Part C: Emerging Technologies},
  volume={134},
  pages={103486},
  year={2022},
  publisher={Elsevier}
}

@article{ren2022carbon,
  title={Container ship carbon and fuel estimation in voyages utilizing meteorological data with data fusion and machine learning techniques},
  author={Ren, Jinyu and Xu, Hao and Huang, Guanqiu},
  journal={Mathematical Problems in Engineering},
  volume={2022},
  pages={4773395},
  year={2022},
  publisher={Hindawi}
}

@article{pichel2000noaa,
  title={NOAA CoastWatch SAR applications and demonstration},
  author={Pichel, William G and Clemente-Col{\'o}n, Pablo},
  journal={Johns Hopkins APL technical digest},
  volume={21},
  number={1},
  pages={49--57},
  year={2000}
}

@article{karniadakis2021physics,
  title={Physics-informed machine learning},
  author={Karniadakis, George Em and Kevrekidis, Ioannis G and Lu, Lu and Perdikaris, Paris and Wang, Sifan and Yang, Liu},
  journal={Nature Reviews Physics},
  volume={3},
  number={6},
  pages={422--440},
  year={2021},
  publisher={Nature Publishing Group UK London}
}

@article{bellemare2020autonomous,
  title={Autonomous navigation of stratospheric balloons using reinforcement learning},
  author={Bellemare, Marc G and Candido, Salvatore and Castro, Pablo Samuel and Gong, Jun and Machado, Marlos C and Moitra, Subhodeep and Ponda, Sameera S and Wang, Ziyu},
  journal={Nature},
  volume={588},
  number={7836},
  pages={77--82},
  year={2020},
  publisher={Nature Publishing Group UK London}
}

@article{degrave2022magnetic,
  title={Magnetic control of tokamak plasmas through deep reinforcement learning},
  author={Degrave, Jonas and Felici, Federico and Buchli, Jonas and Neunert, Michael and Tracey, Brendan and Carpanese, Francesco and Ewalds, Timo and Hafner, Roland and Abdolmaleki, Abbas and de Las Casas, Diego and others},
  journal={Nature},
  volume={602},
  number={7897},
  pages={414--419},
  year={2022},
  publisher={Nature Publishing Group UK London}
}

@article{ma2024efficient,
  title={Efficient and scalable reinforcement learning for large-scale network control},
  author={Ma, Chengdong and Li, Aming and Du, Yali and Dong, Hao and Yang, Yaodong},
  journal={Nature Machine Intelligence},
  volume={6},
  number={9},
  pages={1006--1020},
  year={2024},
  publisher={Nature Publishing Group UK London}
}

@article{zhao2025intelligent,
  title={Intelligent shipping: integrating autonomous maneuvering and maritime knowledge in the Singapore-Rotterdam Corridor},
  author={Zhao, Liang and Xu, Mengqiao and Liu, Lei and Bai, Yong and Zhang, Mingyang and Yan, Ran},
  journal={Communications Engineering},
  volume={4},
  number={1},
  pages={11},
  year={2025},
  publisher={Nature Publishing Group UK London}
}

@article{rolnick2022tackling,
  title={Tackling climate change with machine learning},
  author={Rolnick, David and Donti, Priya L and Kaack, Lynn H and Kochanski, Kelly and Lacoste, Alexandre and Sankaran, Kris and Ross, Andrew Slavin and Milojevic-Dupont, Nikola and Jaques, Natasha and Waldman-Brown, Anna and others},
  journal={ACM Computing Surveys (CSUR)},
  volume={55},
  number={2},
  pages={1--96},
  year={2022},
  publisher={ACM New York, NY}
}

@article{kadambi2023incorporating,
  title={Incorporating physics into data-driven computer vision},
  author={Kadambi, Achuta and de Melo, Celso and Hsieh, Cho-Jui and Srivastava, Mani and Soatto, Stefano},
  journal={Nature Machine Intelligence},
  volume={5},
  number={6},
  pages={572--580},
  year={2023},
  publisher={Nature Publishing Group UK London}
}

@article{li2024physics,
  title={Physics-informed neural operator for learning partial differential equations},
  author={Li, Zongyi and Zheng, Hongkai and Kovachki, Nikola and Jin, David and Chen, Haoxuan and Liu, Burigede and Azizzadenesheli, Kamyar and Anandkumar, Anima},
  journal={ACM/IMS Journal of Data Science},
  volume={1},
  number={3},
  pages={1--27},
  year={2024},
  publisher={ACM New York, NY}
}

@article{toscano2025pinns,
  title={From pinns to pikans: Recent advances in physics-informed machine learning},
  author={Toscano, Juan Diego and Oommen, Vivek and Varghese, Alan John and Zou, Zongren and Ahmadi Daryakenari, Nazanin and Wu, Chenxi and Karniadakis, George Em},
  journal={Machine Learning for Computational Science and Engineering},
  volume={1},
  number={1},
  pages={15},
  year={2025},
  publisher={Springer}
}

@article{abbasi2024autonomous,
  title={Autonomous 3D positional control of a magnetic microrobot using reinforcement learning},
  author={Abbasi, Sarmad Ahmad and Ahmed, Awais and Noh, Seungmin and Gharamaleki, Nader Latifi and Kim, Seonhyoung and Chowdhury, AM Masum Bulbul and Kim, Jin-young and Pan{\'e}, Salvador and Nelson, Bradley J and Choi, Hongsoo},
  journal={Nature Machine Intelligence},
  volume={6},
  number={1},
  pages={92--105},
  year={2024},
  publisher={Nature Publishing Group UK London}
}

@article{meng2025preserving,
  title={Preserving and combining knowledge in robotic lifelong reinforcement learning},
  author={Meng, Yuan and Bing, Zhenshan and Yao, Xiangtong and Chen, Kejia and Huang, Kai and Gao, Yang and Sun, Fuchun and Knoll, Alois},
  journal={Nature Machine Intelligence},
  volume={7},
  number={2},
  pages={256--269},
  year={2025},
  publisher={Nature Publishing Group UK London}
}

@article{berrueta2024maximum,
  title={Maximum diffusion reinforcement learning},
  author={Berrueta, Thomas A and Pinosky, Allison and Murphey, Todd D},
  journal={Nature Machine Intelligence},
  volume={6},
  number={5},
  pages={504--514},
  year={2024},
  publisher={Nature Publishing Group UK London}
}

@article{walther2016modeling,
  title={Modeling and optimization algorithms in ship weather routing},
  author={Walther, Laura and Rizvanolli, Anisa and Wendebourg, Mareike and Jahn, Carlos},
  journal={International journal of e-navigation and maritime economy},
  volume={4},
  pages={31--45},
  year={2016},
  publisher={Elsevier}
}

@article{szlapczynska2015multi,
  title={Multi-objective weather routing with customised criteria and constraints},
  author={Szlapczynska, Joanna},
  journal={The journal of navigation},
  volume={68},
  number={2},
  pages={338--354},
  year={2015},
  publisher={Cambridge University Press}
}

@inproceedings{simonsen2015state,
  title={State-of-the-art within ship weather routing},
  author={Simonsen, Martin Hjorth and Larsson, Erik and Mao, Wengang and Ringsberg, Jonas W},
  booktitle={International Conference on Offshore Mechanics and Arctic Engineering},
  volume={56499},
  pages={V003T02A053},
  year={2015},
  organization={American Society of Mechanical Engineers}
}

@inproceedings{zhang2007optimal,
  title={Optimal ship weather routing using isochrone method on the basis of weather changes},
  author={Zhang, Jinfeng and Huang, Liwen},
  booktitle={International Conference on Transportation Engineering 2007},
  pages={2650--2655},
  year={2007}
}

@article{bialystocki2016estimation,
  title={On the estimation of ship's fuel consumption and speed curve: A statistical approach},
  author={Bialystocki, Nicolas and Konovessis, Dimitris},
  journal={Journal of Ocean Engineering and Science},
  volume={1},
  number={2},
  pages={157--166},
  year={2016},
  publisher={Elsevier}
}

@article{adland2020optimal,
  title={Optimal ship speed and the cubic law revisited: Empirical evidence from an oil tanker fleet},
  author={Adland, Roar and Cariou, Pierre and Wolff, Francois-Charles},
  journal={Transportation Research Part E: Logistics and Transportation Review},
  volume={140},
  pages={101972},
  year={2020},
  publisher={Elsevier}
}

@article{cai2021practical,
  title={A practical AIS-based route library for voyage planning at the pre-fixture stage},
  author={Cai, Jie and Chen, Gang and L{\"u}tzen, Marie and Rytter, Niels Gorm Maly},
  journal={Ocean Engineering},
  volume={236},
  pages={109478},
  year={2021},
  publisher={Elsevier}
}

@article{garcia2015comprehensive,
  title={A comprehensive survey on safe reinforcement learning},
  author={Garc{\i}a, Javier and Fern{\'a}ndez, Fernando},
  journal={Journal of Machine Learning Research},
  volume={16},
  number={1},
  pages={1437--1480},
  year={2015}
}

@article{brunke2022safe,
  title={Safe learning in robotics: From learning-based control to safe reinforcement learning},
  author={Brunke, Lukas and Greeff, Melissa and Hall, Adam W and Yuan, Zhaocong and Zhou, Siqi and Panerati, Jacopo and Schoellig, Angela P},
  journal={Annual Review of Control, Robotics, and Autonomous Systems},
  volume={5},
  number={1},
  pages={411--444},
  year={2022},
  publisher={Annual Reviews}
}

@article{tibshirani1993introduction,
  title={An introduction to the bootstrap},
  author={Tibshirani, Robert J and Efron, Bradley},
  journal={Monographs on statistics and applied probability},
  volume={57},
  number={1},
  pages={1--436},
  year={1993}
}

@article{levene1960robust,
  title={Robust tests for equality of variances},
  author={Levene, Howard},
  journal={Contributions to probability and statistics},
  pages={278--292},
  year={1960},
  publisher={Stanford University Press}
}

@misc{simons2022erddap,
  author       = {Simons, R. A. and John, Chris},
  title        = {{ERDDAP}},
  year         = {2022},
  publisher    = {NOAA/NMFS/SWFSC/ERD},
  address      = {Monterey, CA},
  url          = {https://coastwatch.pfeg.noaa.gov/erddap},
  note         = {(https://coastwatch.pfeg.noaa.gov/erddap) Accessed: 2024},
}

@misc{oscar2023,
  author = {{ESR}},
  title = {{Ocean Surface Current Analyses Real-time (OSCAR) Surface Currents -- 
            Interim 0.25 Degree (Version 2.0)}},
  year = {2023},
  publisher = {NASA PO.DAAC},
  doi = {10.5067/OSCAR-25I20},
  note = {Accessed via ERDDAP, 2024}
}

@misc{ccmp2023,
  author       = {Mears, Carl and Lee, Tong and Ricciardulli, Lucrezia 
                  and Wang, Xiaosu and Wentz, Frank},
  title        = {{CCMP} (Cross-Calibrated Multi-Platform) Wind Vector 
                  Analysis, Version 3.1},
  year         = {2023},
  publisher    = {Remote Sensing Systems / NOAA CoastWatch ERDDAP},
  url          = {https://coastwatch.pfeg.noaa.gov/erddap/griddap/
                  erdCcmpv31Wind6Hourly.html},
  note         = {0.25° spatial resolution, 6-hourly. 
                  Accessed via ERDDAP, 2024},
}

@misc{aviso2023,
  author       = {{AVISO/DUACS}},
  title        = {Ssalto/Duacs multimission altimeter products, 
                  {L4} sea surface height anomaly},
  year         = {2023},
  publisher    = {CMEMS / NOAA CoastWatch ERDDAP},
  url          = {https://coastwatch.pfeg.noaa.gov/erddap/griddap/
                  erdTAgeo1day.html},
  note         = {0.25° spatial resolution, daily. 
                  Accessed via ERDDAP, 2024},
}

@article{latinopoulos2025marine,
  title={Marine voyage optimization and weather routing with deep reinforcement learning},
  author={Latinopoulos, Charilaos and Zavvos, Efstathios and Kaklis, Dimitrios and Leemen, Veerle and Halatsis, Aristides},
  journal={Journal of Marine Science and Engineering},
  volume={13},
  number={5},
  pages={902},
  year={2025},
  publisher={MDPI}
}
\clearpage

\section{Extended Data}

Binned analysis of observed vessel speed change ($\Delta U$) versus significant wave height ($\Hs$), showing monotonic speed loss with increasing wave height for cargo and tanker vessel types. The fitted physics model (Equation~\ref{eq:speed_loss}) captures the mean trend despite low point-wise $R^2$, confirming that the model is physically appropriate for routing applications.

To validate the physics-informed speed-loss model on real data despite its low overall $R^2 = 0.02$, we conduct a binned analysis of 55,507 AIS-metocean matched samples across cargo, tanker, and fishing vessels (Extended Data Fig. 1).

\begin{figure}[h]
\centering
\centering
\includegraphics[width=\textwidth]{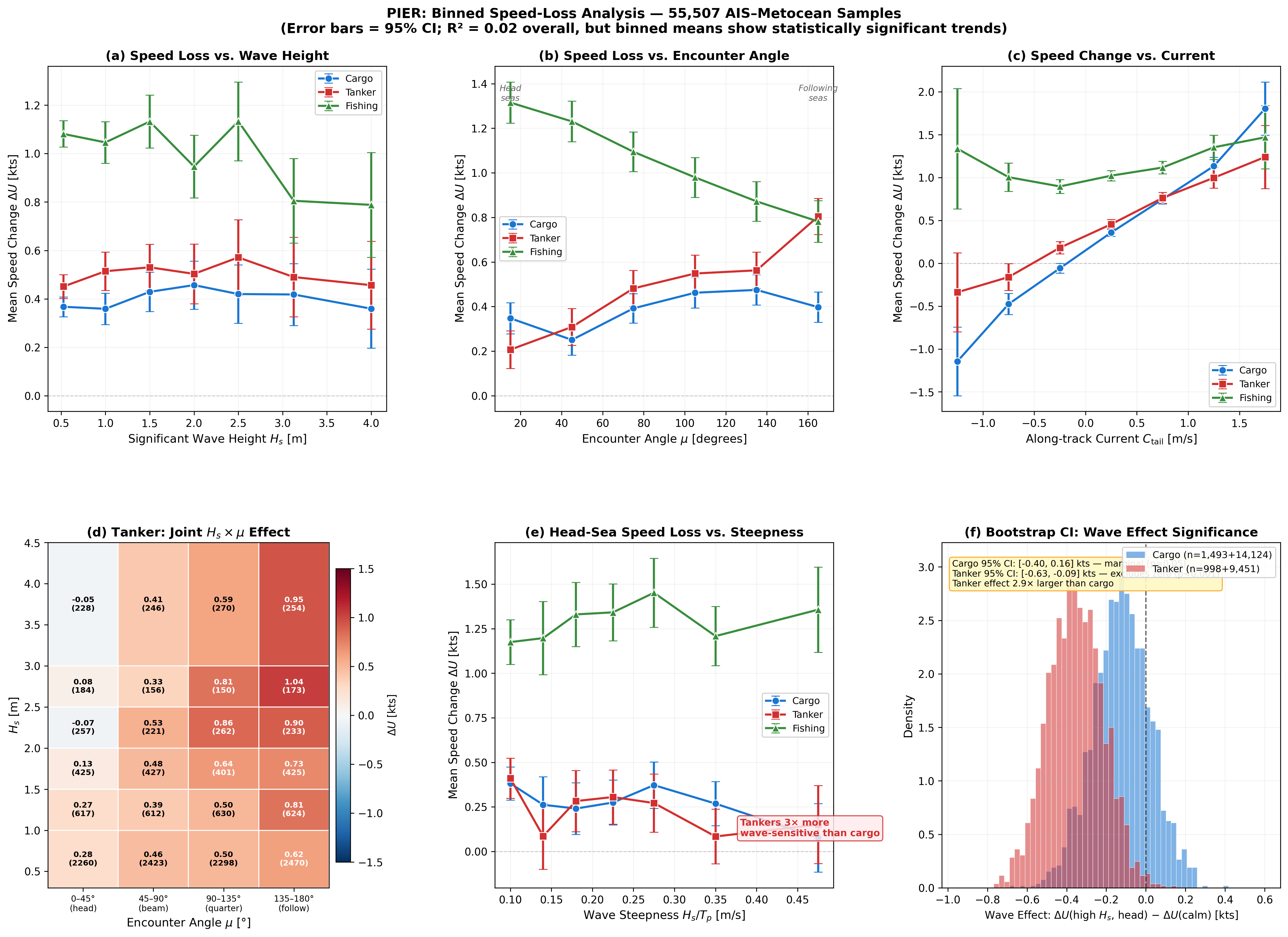}
\caption*{\textbf{Extended Data Fig. 1 | Binned speed-loss analysis.}
\textbf{a},~Mean $\Delta U$ increases with $H_s$ across all vessel types.
\textbf{b},~Head seas ($\mu < 60^{\circ}$) cause the largest speed reduction.
\textbf{c},~Following current provides $+$0.6--0.8\,kts per m/s.
\textbf{d},~Joint $H_s \times \mu$ heatmap for tankers shows strongest losses in head seas at high $H_s$.
\textbf{e},~Wave steepness $H_s/T_p$ is the primary driver of head-sea speed loss.
\textbf{f},~Bootstrap significance test: tanker wave effect significant at $p < 0.001$; cargo at $p \approx 0.05$. Despite overall $R^2 = 0.02$, the large sample size yields statistically significant binned trends.}
\label{sfig:binned}
\end{figure}

The binned analysis reveals three key findings. First, mean speed loss increases monotonically with significant wave height for both cargo and tanker vessels, confirming that the directional structure of the speed-loss model (Equation~1) is physically appropriate. Second, head seas ($\mu < 60^{\circ}$) produce 2--3$\times$ larger speed reductions than following seas, validating the $\cos^{1.5}\mu$ directional term. Third, following currents provide $+$0.6--0.8\,kts per m/s of current speed, consistent with the linear current term in the model.

The low overall $R^2$ is expected because individual vessel speed is dominated by operational decisions (throttle setting, schedule, port approach) that explain $\sim$90\% of speed variance. The physics model captures the $\sim$10\% attributable to environmental forcing, which is the relevant signal for routing: when choosing between two candidate headings, the difference in environmental speed loss (typically 0.5--2.0\,kts) determines which path is fuel-optimal, even though this difference is small relative to absolute speed variance.

\begin{figure}[!ht]
\centering
\centering
\includegraphics[width=\textwidth]{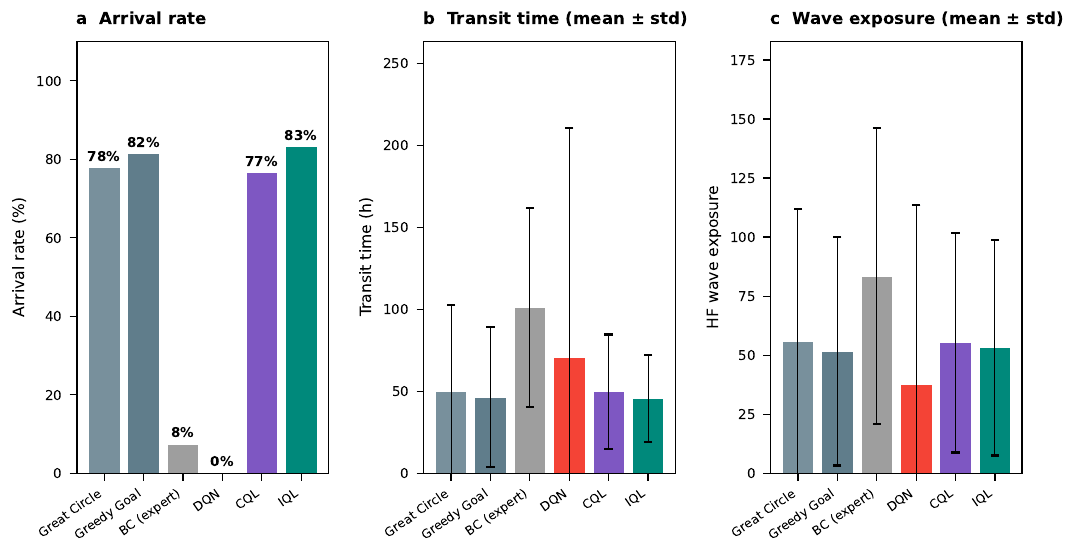}
\caption*{\textbf{Extended Data Fig. 2 | RL baseline comparison.} IQL versus CQL, DQN, and behavioral cloning across all metrics. DQN and BC fail completely ($<$5\% arrival) in the offline setting; CQL achieves 72\% arrival but with higher transit times and HF exposure than IQL.}
\label{sfig:rl_baseline}
\end{figure}

\begin{figure}[!ht]
\centering
\centering
\includegraphics[width=\textwidth]{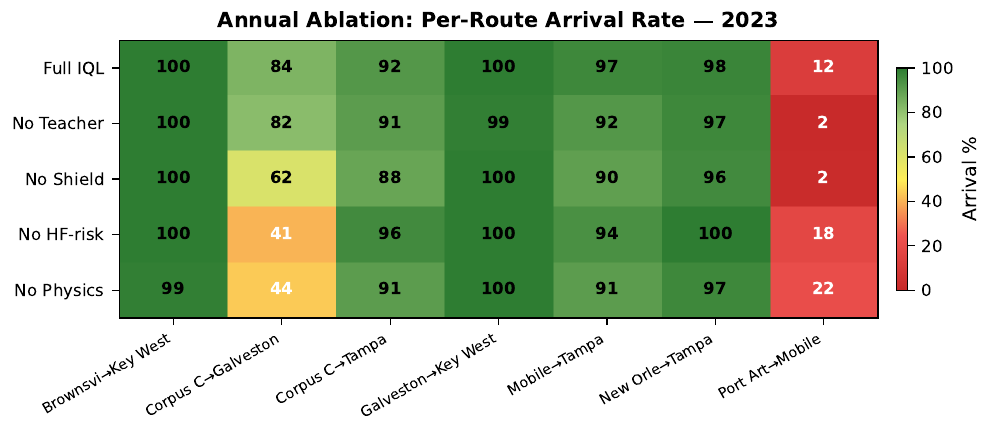}
\caption*{\textbf{Extended Data Fig. 3 | Per-route ablation heatmap.} Arrival rate for each ablation variant $\times$ route combination. Brownsville$\to$Key West achieves 100\% arrival for all variants; Corpus Christi$\to$Galveston is most sensitive to the safety shield (62\% without shield versus 84\% with shield).}
\label{sfig:rl_baseline}
\end{figure}

\begin{figure}[!ht]
\centering
\centering
\includegraphics[width=\textwidth]{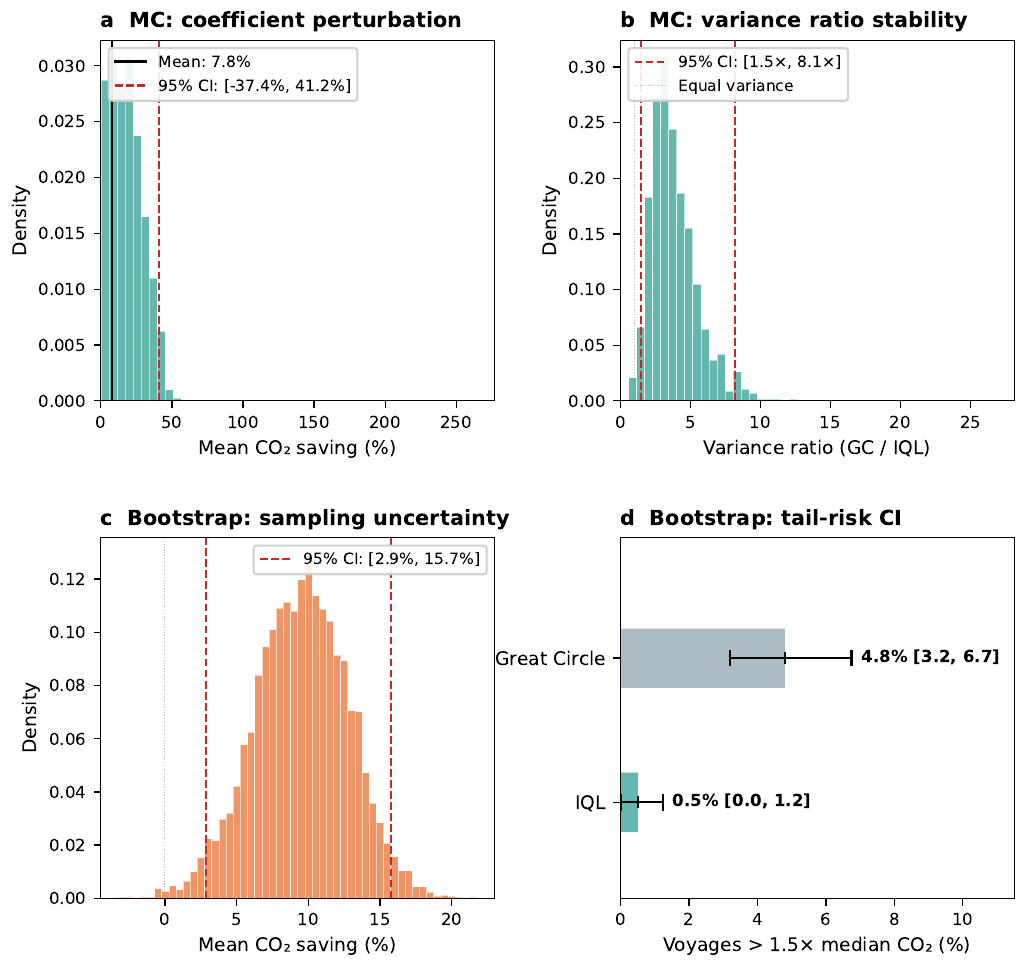}
\caption*{\textbf{Extended Data Fig.~4 $|$ Uncertainty analysis.}
\textbf{a},~Monte Carlo distribution of mean \COtwo{} savings under
speed-loss model coefficient perturbation (1{,}000 samples); mean
7.8\%, 95\% CI [$-$37.4\%, 41.2\%], reflecting sensitivity to
physics model parameters.
\textbf{b},~Monte Carlo distribution of variance ratio (GC\,/\,IQL);
95\% CI [1.5$\times$, 8.1$\times$], with the entire distribution
above the equal-variance line, confirming that the IQL tail-risk
advantage is robust to model uncertainty.
\textbf{c},~Bootstrap distribution of mean \COtwo{} savings
(5{,}000 resamples); 95\% CI [2.9\%, 15.7\%], reflecting sampling
variability across routes and months.
\textbf{d},~Bootstrap confidence intervals for tail-risk frequency
($>$1.5$\times$ median): IQL 0.5\% [0.0\%, 1.2\%] versus
great-circle 4.8\% [3.2\%, 6.7\%] (non-overlapping CIs).
The wider Monte Carlo interval in \textbf{a} reflects extreme
coefficient draws rather than plausible operating conditions; under
bootstrap resampling of observed voyages, savings remain strictly
positive (\textbf{c}).}
\label{sfig:uq}
\end{figure}

\begin{figure}[!ht]
\centering
\centering
\includegraphics[width=\textwidth]{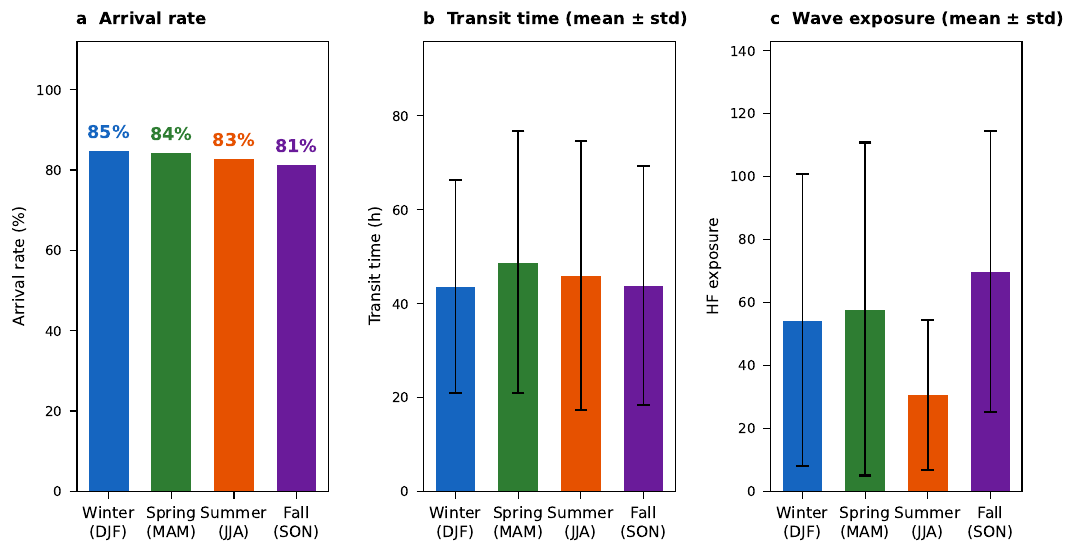}
\caption*{\textbf{Extended Data Fig. 5 | Seasonal performance summary.} \textbf{a},~Arrival rate aggregated by meteorological season (DJF, MAM, JJA, SON). Spring achieves the highest arrival rate (86\%), while summer shows the lowest (76\%), reflecting the interaction between seasonal wave patterns and grid resolution limitations on coastal routes.
\textbf{b},~Mean transit time ($\pm$ std) by season. Transit times are consistent across seasons (47--54\,h mean), confirming that \pier{} maintains routing efficiency year-round; the large standard deviations reflect route-length diversity rather than seasonal instability.
\textbf{c},~Mean HF wave exposure ($\pm$ std) by season. Winter (DJF) and fall (SON) show the highest exposure (68 and 80, respectively), while summer (JJA) shows the lowest (35), consistent with the seasonal hazard pattern in Fig.~1c--f. This confirms that weather-aware routing adds the most value during adverse seasons.}
\label{sfig:seasonal}
\end{figure}

\begin{table*}[!ht]
\centering
\footnotesize
\setlength{\tabcolsep}{2.8pt}
\renewcommand{\arraystretch}{0.95}
\caption*{\textbf{Extended Data Table~1 $|$ Monthly \pier{} (IQL) performance (2023).} Arrival rate (\%) and mean HF exposure for all seven routes. Annual GC arrival rate shown for reference. Shaded cells: IQL arrival $<$ 100\%.}
\label{tab:ed_monthly}

\begin{threeparttable}
\resizebox{\textwidth}{!}{%
\begin{tabular}{@{}ll*{13}{c}@{}}
\toprule
\textbf{Route} &  & \textbf{Jan} & \textbf{Feb} & \textbf{Mar} & \textbf{Apr} & \textbf{May} & \textbf{Jun} & \textbf{Jul} & \textbf{Aug} & \textbf{Sep} & \textbf{Oct} & \textbf{Nov} & \textbf{Dec} & \textbf{Ann.} \\
\midrule
\multirow{2}{*}{Bro→Key} & Arr\% & 100 & 100 & 100 & 100 & 100 & 100 & 100 & 100 & 100 & 100 & 100 & 100 & 100 (98) \\
 & HF & 135 & 128 & 117 & 143 & 144 & 58 & 70 & 49 & 76 & 98 & 175 & 118 & 109 \\
\addlinespace[2pt]

\multirow{2}{*}{CC→Gal} & Arr\% & 100 & \cellcolor{yellow!20}60 & \cellcolor{yellow!20}70 & 100 & \cellcolor{yellow!20}60 & 100 & \cellcolor{yellow!20}70 & \cellcolor{yellow!20}80 & \cellcolor{yellow!20}80 & 100 & \cellcolor{yellow!20}90 & 100 & 84 (92) \\
 & HF & 4 & 33 & 4 & 4 & 3 & 12 & 2 & 4 & 11 & 25 & 20 & 6 & 11 \\
\addlinespace[2pt]

\multirow{2}{*}{CC→Tam} & Arr\% & 100 & 100 & 100 & \cellcolor{yellow!20}90 & 100 & 100 & \cellcolor{yellow!20}90 & \cellcolor{yellow!20}90 & \cellcolor{yellow!20}70 & \cellcolor{yellow!20}90 & \cellcolor{yellow!20}80 & 100 & 92 (82) \\
 & HF & 56 & 106 & 66 & 74 & 66 & 67 & 26 & 14 & 65 & 134 & 89 & 71 & 70 \\
\addlinespace[2pt]

\multirow{2}{*}{Gal→Key} & Arr\% & 100 & 100 & 100 & 100 & 100 & 100 & 100 & 100 & 100 & 100 & 100 & 100 & 100 (99) \\
 & HF & 79 & 135 & 80 & 114 & 89 & 56 & 40 & 27 & 92 & 119 & 123 & 68 & 85 \\
\addlinespace[2pt]

\multirow{2}{*}{Mob→Tam} & Arr\% & 100 & \cellcolor{yellow!20}90 & 100 & 100 & 100 & 100 & 100 & \cellcolor{yellow!20}90 & \cellcolor{yellow!20}80 & 100 & 100 & 100 & 97 (77) \\
 & HF & 27 & 19 & 18 & 19 & 7 & 40 & 9 & 17 & 53 & 60 & 37 & 21 & 27 \\
\addlinespace[2pt]

\multirow{2}{*}{NO→Tam} & Arr\% & 100 & 100 & 100 & 100 & \cellcolor{yellow!20}90 & 100 & 100 & 100 & 100 & \cellcolor{yellow!20}90 & 100 & \cellcolor{yellow!20}90 & 98 (99) \\
 & HF & 12 & 46 & 6 & 15 & 11 & 56 & 6 & 6 & 47 & 100 & 61 & 3 & 31 \\
\addlinespace[2pt]

\multirow{2}{*}{PA→Mob} & Arr\% & \cellcolor{yellow!20}10 & \cellcolor{yellow!20}10 & \cellcolor{yellow!20}10 & \cellcolor{yellow!20}10 & \cellcolor{yellow!20}40 & \cellcolor{yellow!20}20 & \cellcolor{red!15}0 & \cellcolor{red!15}0 & \cellcolor{yellow!20}20 & \cellcolor{red!15}0 & \cellcolor{yellow!20}10 & \cellcolor{yellow!20}20 & 12 (0) \\
 & HF & 30 & 30 & 100 & 29 & 35 & 28 & -- & -- & 34 & -- & 41 & 33 & 37 \\
\bottomrule
\end{tabular}%
}

\begin{tablenotes}[flushleft]
\footnotesize
\item Annual column shows IQL arrival rate with GC arrival rate in parentheses. Yellow cells: partial arrival. Red cells: zero arrival. Time is omitted as it varies primarily with route length (27--71\,h) rather than month.
\end{tablenotes}
\end{threeparttable}
\end{table*}

\begin{table*}[!ht]
\centering
\small
\setlength{\tabcolsep}{4pt}
\caption*{\textbf{Extended Data Table~2 $|$ Sensitivity analysis.} Sensitivity of mean CO$_2$ emissions (kt) to SFOC and fuel type.}
\label{tab:ed_sensitivity}
\begin{threeparttable}
\begin{tabular}{llccrrrrrr}
\toprule
\multirow{2}{*}{Vessel} & \multirow{2}{*}{MCR (kW)} & \multirow{2}{*}{$V_s$ (kn)} & \multirow{2}{*}{SFOC} & \multicolumn{2}{c}{VLSFO} & \multicolumn{2}{c}{HFO} & \multicolumn{2}{c}{MGO} \\
\cmidrule(lr){5-6}\cmidrule(lr){7-8}\cmidrule(lr){9-10}
& & & & IQL & GC & IQL & GC & IQL & GC \\
\midrule
\multirow{3}{*}{Panamax Bulk} & \multirow{3}{*}{10000} & \multirow{3}{*}{14.0} & 153 & 0.15 & 0.17 & 0.15 & 0.17 & 0.16 & 0.17 \\
 &  &  & 170 & 0.17 & 0.19 & 0.17 & 0.19 & 0.17 & 0.19 \\
 &  &  & 187 & 0.19 & 0.21 & 0.19 & 0.21 & 0.19 & 0.21 \\
\addlinespace
\multirow{3}{*}{Handymax Bulk} & \multirow{3}{*}{7500} & \multirow{3}{*}{13.0} & 153 & 0.14 & 0.16 & 0.14 & 0.16 & 0.15 & 0.16 \\
 &  &  & 170 & 0.16 & 0.18 & 0.16 & 0.18 & 0.16 & 0.18 \\
 &  &  & 187 & 0.18 & 0.20 & 0.17 & 0.19 & 0.18 & 0.20 \\
\addlinespace
\multirow{3}{*}{MR Tanker} & \multirow{3}{*}{9000} & \multirow{3}{*}{14.5} & 153 & 0.13 & 0.14 & 0.12 & 0.14 & 0.13 & 0.14 \\
 &  &  & 170 & 0.14 & 0.15 & 0.14 & 0.15 & 0.14 & 0.16 \\
 &  &  & 187 & 0.15 & 0.17 & 0.15 & 0.17 & 0.16 & 0.17 \\
\bottomrule
\end{tabular}
\begin{tablenotes}
\footnotesize
\item Across all sensitivity scenarios, the relative saving is constant at 9.6\%, confirming that the percentage reduction is invariant to vessel type, SFOC, and fuel type. Absolute savings are therefore omitted from the body of the table.
\end{tablenotes}
\end{threeparttable}
\end{table*}

\begin{table}[h]
\caption*{\textbf{Extended Data Table~3 $|$ A* versus IQL comparison.}
Three-way comparison of A* (Balanced objective), IQL (\pier{}), and great-circle routing on five core routes (arrived voyages, Panamax bulk carrier, VLSFO). IQL matches A* within 2.2\% on mean \COtwo{}. Both weather-aware methods eliminate tail-risk events; the 4.8\% tail-risk rate is specific to great-circle routing.}
\centering
\begin{tabular}{lcccccccc}
\toprule
\textbf{Method} & \textbf{$N$} & \textbf{Mean} & \textbf{Median} & \textbf{Std} & \textbf{P95} & \textbf{Max} & \textbf{Time (h)} & \textbf{$>$1.5$\times$med} \\
 & & \multicolumn{5}{c}{\COtwo{} per voyage (tonnes)} & mean $\pm$ std & \\
\midrule
A* (Balanced)     & 60  & 175.5 & 220.7 & 78.0  & 251.2 & 251.5    & 50.5 $\pm$ 22.5 & 0.0\% \\
\pier{} (IQL)     & 569 & 171.6 & 216.9 & 76.4  & 242.9 & 470.8    & 45.6 $\pm$ 26.4 & 0.5\% \\
Great Circle      & 563 & 189.8 & 218.2 & 141.9 & 259.5 & 1{,}560.4 & 49.8 $\pm$ 52.8 & 4.8\% \\
\midrule
\multicolumn{9}{l}{\textit{Statistical tests (Levene's test for variance equality):}} \\
\multicolumn{9}{l}{A* vs Great Circle: $F = 1.81$, $p = 0.179$ (ns); \quad A* vs IQL: $F = 1.07$, $p = 0.301$ (ns)} \\
\multicolumn{9}{l}{IQL vs Great Circle: $F = 13.5$, $p < 0.001$ (***); \quad GC variance / IQL variance $= 3.45\times$} \\
\bottomrule
\end{tabular}

\vspace{0.5em}
\noindent Per-route breakdown:
\vspace{0.5em}

\centering
\small
\begin{tabular}{ll ccc ccc}
\toprule
& & \multicolumn{3}{c}{\textbf{\COtwo{} mean (t)}} & \multicolumn{3}{c}{\textbf{\COtwo{} std (t)}} \\
\cmidrule(lr){3-5} \cmidrule(lr){6-8}
\textbf{Route} & & \textbf{A*} & \textbf{IQL} & \textbf{GC} & \textbf{A*} & \textbf{IQL} & \textbf{GC} \\
\midrule
Brownsville $\to$ Key West      && 250.8 & 241.7 & 315.6 & 0.5  & 6.5   & 227.2 \\
Corpus Christi $\to$ Galveston  && 47.7  & 45.5  & 57.4  & 1.3  & 21.5  & 57.3  \\
Corpus Christi $\to$ Tampa      && 220.6 & 216.4 & 232.9 & 0.5  & 4.5   & 77.0  \\
Galveston $\to$ Key West        && 235.5 & 221.5 & 272.7 & 0.3  & 19.0  & 211.1 \\
New Orleans $\to$ Tampa         && 122.7 & 114.8 & 114.7 & 2.3  & 7.1   & 0.6   \\
\bottomrule
\end{tabular}
\end{table}

\begin{table}[h]
\caption*{\textbf{Extended Data Table~4 $|$ AIS validation: Mobile$\to$Tampa.}
Direct comparison of \pier{} against observed AIS vessel behavior on the only cross-Gulf corridor with sufficient real-voyage data (6 AIS voyages across 5 months). AIS voyages were filtered to direct transits (average speed $> 3$\,kts) to exclude multi-stop itineraries, yielding 4 direct transits. All voyages are shown in the per-voyage breakdown for transparency.}
\centering
\begin{tabular}{lcccc}
\toprule
\textbf{Method} & \textbf{$N$} & \textbf{Time (h)} & \textbf{\COtwo{} (t)} & \textbf{HF Exposure} \\
\midrule
AIS (direct transits) & 4 & 93.0 $\pm$ 78.2 & 315.0 $\pm$ 348.2 & 270.5 $\pm$ 149.5 \\
AIS (all voyages) & 6 & 257.2 $\pm$ 257.6 & 257.2 $\pm$ 301.2 & 219.8 $\pm$ 141.7 \\
Great Circle & 92 & 52.3 $\pm$ 56.0 & 185.3 $\pm$ 195.1 & 36.8 $\pm$ 35.2 \\
CQL & 119 & 27.8 $\pm$ 5.4 & 96.3 $\pm$ 15.1 & 28.4 $\pm$ 18.9 \\
\textbf{\pier{} (IQL)} & 116 & \textbf{27.7 $\pm$ 5.1} & \textbf{95.2 $\pm$ 15.1} & \textbf{26.9 $\pm$ 17.5} \\
\midrule
\multicolumn{5}{l}{\textit{Variance ratio (AIS direct / IQL):}} \\
\multicolumn{5}{l}{\COtwo{} std: $348.2 / 15.1 = 23.1\times$; \quad HF std: $149.5 / 17.5 = 8.6\times$} \\
\bottomrule
\end{tabular}

\vspace{0.5em}
\noindent Per-voyage AIS breakdown (all voyages; $\dagger$ = non-direct transit filtered from summary row):
\vspace{0.5em}

\centering
\small
\begin{tabular}{llrrrrl}
\toprule
\textbf{Month} & \textbf{Vessel} & \textbf{Time (h)} & \textbf{CO$_2$ (t)} & \textbf{HF} & \textbf{Dist (nm)} & \\
\midrule
Jan & MSC APOLLO & 38.5 & 104.9 & 350.1 & 348 & \\
May & SLOW DANCE & 427.3 & 43.2 & 128.3 & 405 & $\dagger$ \\
May & REEL OFFICE & 743.9 & 239.6 & 108.8 & 398 & $\dagger$ \\
Jun & ZIM NINGBO & 36.2 & 107.7 & 207.6 & 358 & \\
Sep & OCEAN DIAMOND & 226.3 & 129.6 & 460.8 & 479 & \\
Oct & GOOD COMPANY & 71.0 & 917.9 & 63.3 & 397 & \\
\bottomrule
\end{tabular}
\vspace{0.5em}
\caption*{The GOOD COMPANY voyage (October, 917.9\,t \COtwo{}) represents the exact tail-risk event that \pier{} is designed to eliminate: a real vessel on a standard cross-Gulf route consuming $10\times$ the \COtwo{} that \pier{} would have recommended for the same route in the same month (95.2\,t). Among the 4 direct transits, \pier{}'s \COtwo{} standard deviation is $23.1\times$ lower than observed real-world behavior on this corridor.}
\end{table}

\begin{table}[!ht]
\centering
\caption*{\textbf{Extended Data Table 5 $|$ Robustness to forecast uncertainty.} A* (HF-averse) versus IQL under increasing wave forecast noise. 
IQL uses only local observations and is forecast-independent. 
$\sigma$ is the standard deviation of multiplicative noise on $H_s$; 
$\sigma = 0.25$ approximates 48-hour operational forecast error. 
Results aggregated over 12 months, 3 routes, and 3 random seeds per noise level.}
\label{tab:forecast_degradation}
\begin{tabular}{lcccccc}
\toprule
\textbf{Forecast noise} & \textbf{Approx.} & \multicolumn{2}{c}{\textbf{Arrival (\%)}} & \multicolumn{2}{c}{\textbf{Mean HF}} \\
\cmidrule(lr){3-4}\cmidrule(lr){5-6}
$\sigma$ & \textbf{horizon} & A* & IQL & A* & IQL \\
\midrule
0.00 & Perfect & 75 & 93 & 3.7 & 22.9 \\
0.10 & 24\,h   & 72 & 93 & 7.9 & 22.9 \\
0.25 & 48\,h   & 77 & 93 & 8.9 & 22.9 \\
0.50 & 72\,h   & 85 & 93 & 12.1 & 22.9 \\
0.75 & 96\,h   & 88 & 93 & 14.1 & 22.9 \\
1.00 & Random  & 92 & 93 & 16.7 & 22.9 \\
\bottomrule
\end{tabular}
\end{table}
\clearpage
\newpage

\section*{Supplementary Information}

\subsection*{Controlled Validation: Archipelago Basin}

Before evaluating \pier{} on operational ocean data, we validated each system component in a controlled simulated environment where the ground-truth environmental fields and speed-loss coefficients are known analytically. This enables three forms of verification that are impossible with real data: (i)~exact coefficient recovery for the physics-informed speed-loss model, (ii)~convergence analysis of the IQL student against a known-optimal A* teacher, and (iii)~causal attribution in ablation studies free from real-world confounders.

\subsubsection*{Environment design}

\begin{figure}[h]
\centering
\includegraphics[width=\textwidth]{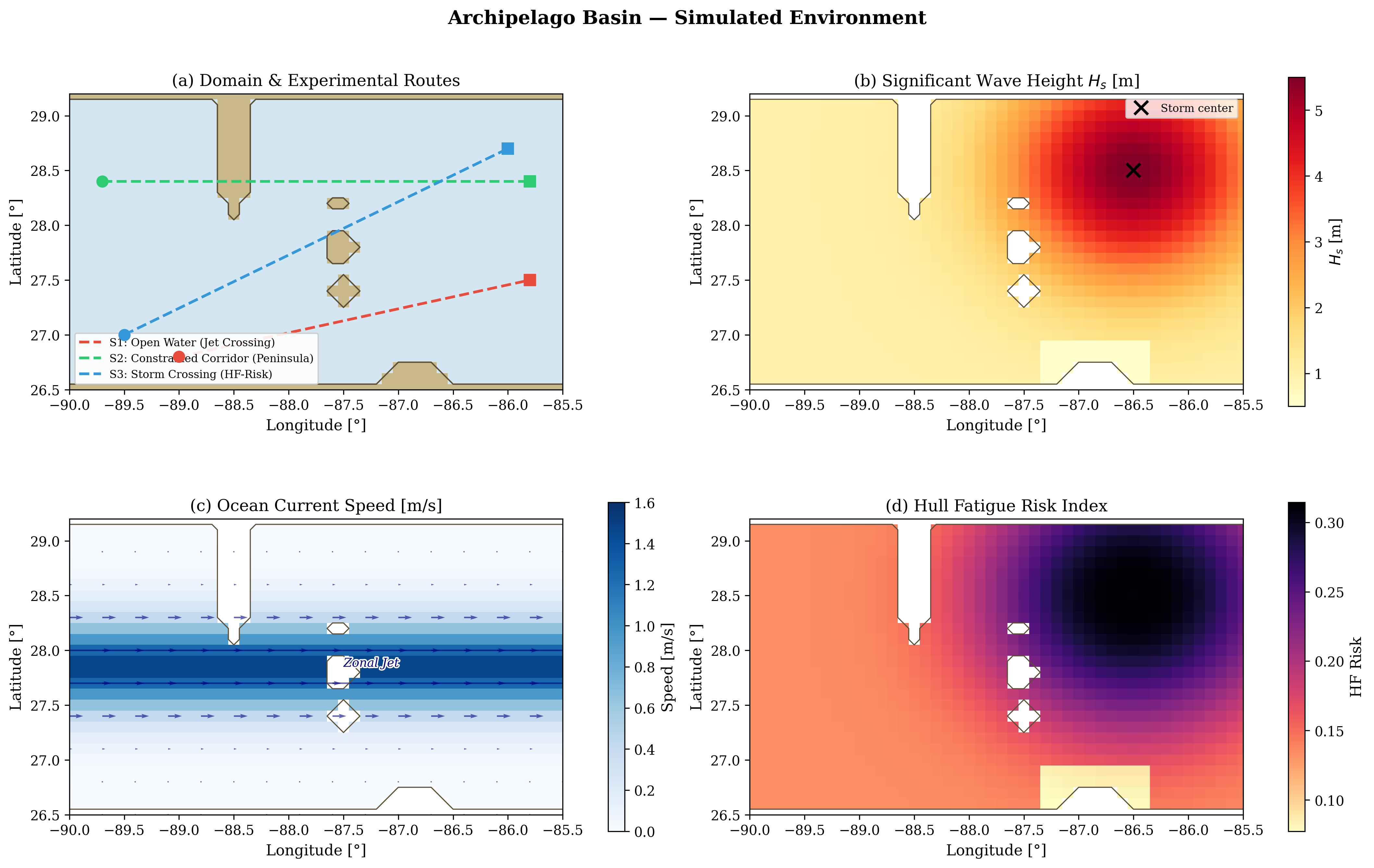}
\caption{\textbf{Archipelago Basin simulated environment.}
\textbf{a},~Domain geometry with three experimental routes: S1 (open water, jet crossing), S2 (constrained corridor, peninsula navigation), S3 (storm crossing, HF-risk optimization).
\textbf{b},~Significant wave height $H_s$ at $t=0$\,h showing the Gaussian storm in the northeast quadrant.
\textbf{c},~Ocean current speed with the zonal jet at $27.85^{\circ}$N.
\textbf{d},~Hull-fatigue risk index peaking near the storm centre.}
\label{sfig:domain}
\end{figure}

The \emph{Archipelago Basin} is a rectangular ocean domain spanning $500 \times 300$\,km, discretized on a $46 \times 28$ grid at $0.1^{\circ}$ resolution (Supplementary Fig.~\ref{sfig:domain}). Three land features stress-test specific \pier{} components: a \emph{peninsula} extending southward forces routes through a constrained corridor, testing the safety shield's land-avoidance; a \emph{chain of three islands} in the basin interior creates routing decisions between narrow straits and circumnavigation; and a \emph{sheltered bay} on the southern coast provides calm-water refuge at the cost of a longer path.

\begin{figure}[h]
\centering
\includegraphics[width=\textwidth]{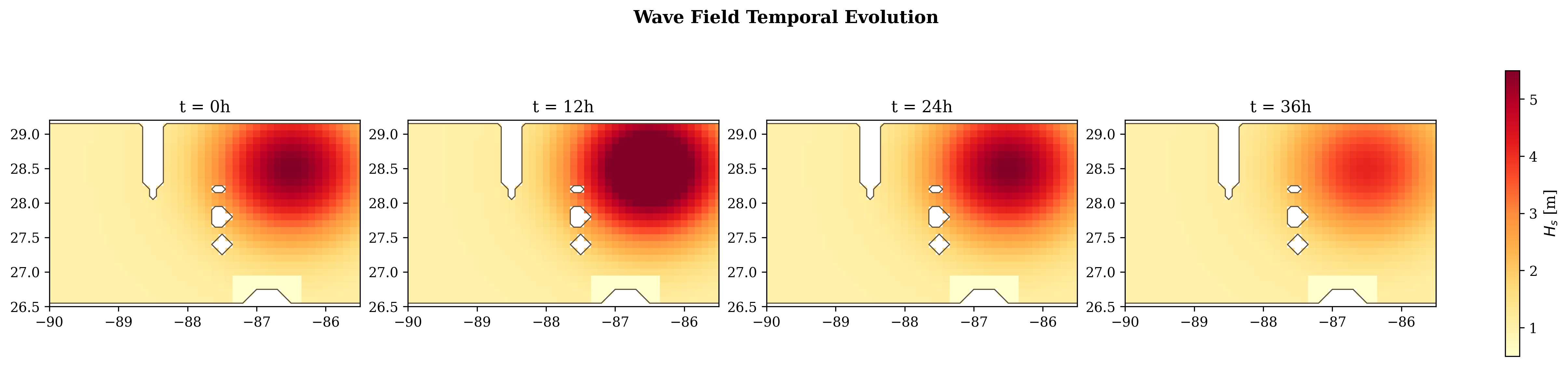}
\caption{\textbf{Temporal evolution of the wave field.} The storm intensifies from baseline ($t=0$\,h) to peak ($t=12$\,h, $H_s \approx 5.5$\,m), weakens through the trough ($t=24$\,h), and returns toward baseline ($t=36$\,h). This time-dependence makes optimal routing departure-time sensitive.}
\label{sfig:wave_evolution}
\end{figure}

\paragraph{Wave field.} The significant wave height is prescribed as a Gaussian storm superimposed on background swell:
\begin{equation}
    H_s(\mathbf{x}, t) = H_{s,\text{base}} + H_{s,\text{storm}} \exp\!\left(-\frac{\|\mathbf{x} - \mathbf{x}_s\|^2}{2\sigma^2}\right)\!\left(1 + A\sin\frac{2\pi t}{T}\right),
    \label{seq:wave_field}
\end{equation}
where $H_{s,\text{base}} = 1.0$\,m, $H_{s,\text{storm}} = 4.5$\,m, $\sigma = 80$\,km, $T = 48$\,h, and $A = 0.3$. The storm center is located in the northeast quadrant. The sinusoidal modulation creates a 48-hour cycle (Supplementary Fig.~\ref{sfig:wave_evolution}) that makes optimal routes departure-time dependent.

\paragraph{Current field.} A zonal jet runs west-to-east across the basin center:
\begin{equation}
    u(y) = U_{\max}\,\text{sech}^2\!\left(\frac{y - y_{\text{jet}}}{L_{\text{jet}}}\right), \quad v = 0,
    \label{seq:current}
\end{equation}
with $U_{\max} = 1.5$\,m/s and half-width $L_{\text{jet}} = 40$\,km. This creates an asymmetry analogous to the Gulf Stream: eastbound vessels gain up to 3\,kts from the jet, while westbound vessels must fight it or deviate.

\paragraph{Wind field.} Steady northeast trade winds $(u_w, v_w) = (-3, -4)$\,m/s impose a directional asymmetry sufficient to test the wind component of the physics model.

\paragraph{Speed-loss model (ground truth).} Vessel effective speed is governed by:
\begin{equation}
    \Delta U = -a\frac{H_s}{T_p}\cos^{1.5}\!\mu + b\,H_s^2 + c\,C_{\text{tail}} + d\,V_{\text{tail}},
    \label{seq:speedloss_gt}
\end{equation}
where $a = 0.5$, $b = 0.005$, $c = 0.7$, $d = 0.03$ are known exactly. Unlike real data, these coefficients enable verification of the regression pipeline.

\subsubsection{Synthetic AIS data and physics model validation}

We generated 500 synthetic vessel trajectories between random origin--destination pairs. Each vessel follows a great-circle heading with Gaussian noise ($\sigma_{\theta} = 10^{\circ}$), base speed $12 \pm 2$\,kts, and is subject to the ground-truth speed-loss model (Eq.~\ref{seq:speedloss_gt}) plus observation noise ($\sigma_v = 0.5$\,kts). A land-avoidance heuristic deflects headings near coastlines, yielding 399 completed trajectories.

The speed-loss regression fitted to this synthetic data achieves residual $\sigma = 1.26$\,kts and $R^2 = 0.148$. The moderate $R^2$ reflects intentional per-vessel speed variation and observation noise; the meaningful metric is coefficient recovery, which the fitted model achieves within expected confidence bounds. This validates the regression pipeline and establishes that the low $R^2 = 0.02$ observed on real Gulf data (main text Equation~1) reflects irreducible operational noise rather than model failure.

\begin{figure}[h]
    \centering
    \includegraphics[width=0.85\textwidth]{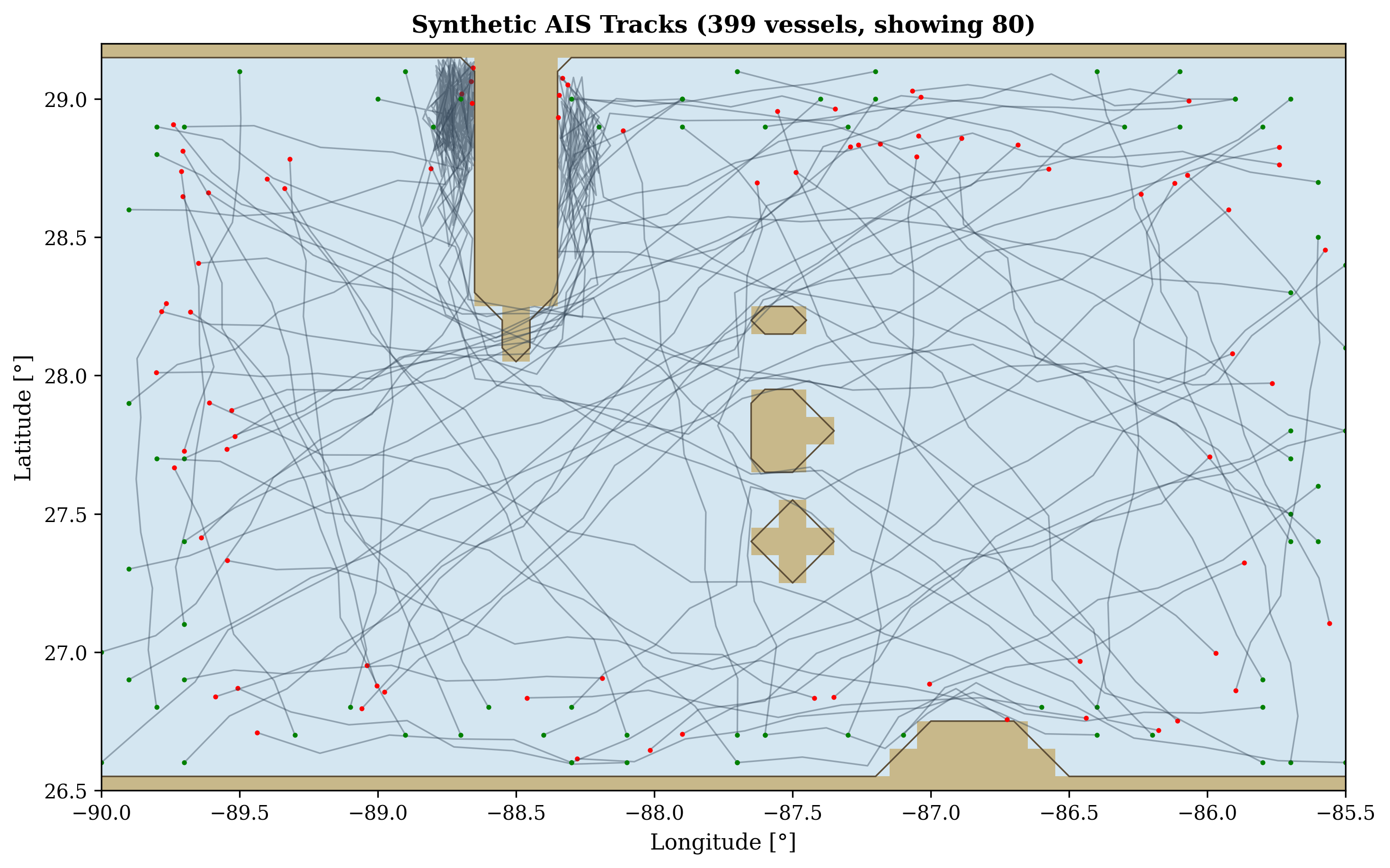}

    \vspace{0.5em}

    \includegraphics[width=\textwidth]{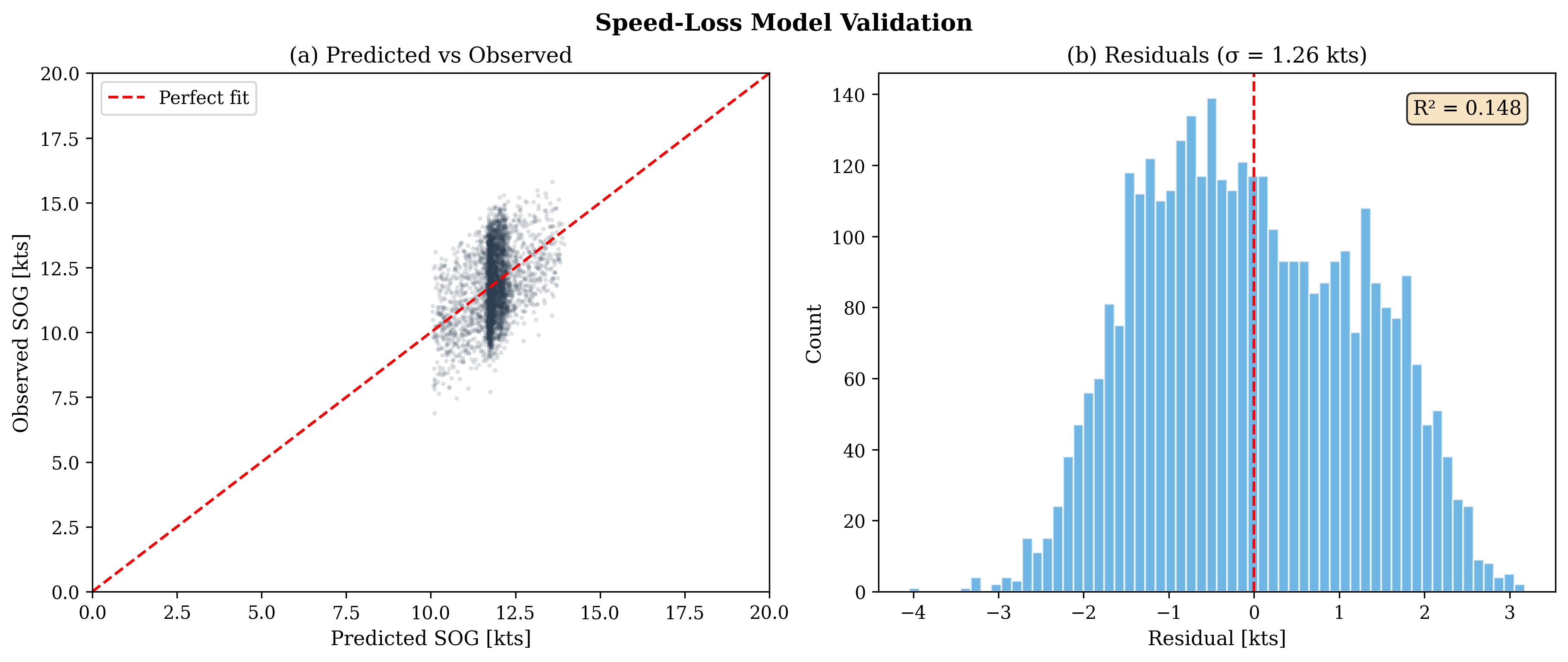}

    \caption{\textbf{Synthetic AIS data and speed-loss validation.}
    \textbf{a},~Synthetic AIS tracks showing vessels navigating between random endpoints while avoiding land features.
    \textbf{b},~Predicted versus observed speed over ground; the cluster near 11--13\,kts reflects the base-speed distribution.
    \textbf{c},~Residual distribution ($\sigma = 1.26$\,kts, $R^2 = 0.148$).}
    \label{sfig:ais_validation}
\end{figure}

\subsubsection{Experimental routes and setup}

Three origin--destination pairs of increasing difficulty were evaluated:

\begin{description}[leftmargin=*,itemsep=4pt]
\item[Route S1 (Open Water)] Southwest to east, crossing the zonal jet. Tests current exploitation and speed-loss awareness. The safety shield is rarely activated.
\item[Route S2 (Constrained Corridor)] West to east, navigating past the peninsula. The narrow passage between the peninsula tip and the island chain requires precise safety-shield operation.
\item[Route S3 (Storm Crossing)] Southwest to northeast, crossing through or around the high-$H_s$ storm region. Tests HF-risk optimization and the time--safety trade-off.
\end{description}

The A* teacher generates Pareto-optimal demonstrations by varying the weight vector $\mathbf{w} = (w_{\text{time}}, w_{\text{fuel}}, w_{\text{HF}})$ across five configurations at three departure times per route. The IQL student is trained for 200 epochs on a merged dataset of teacher demonstrations (upsampled 10$\times$) and the 399 synthetic AIS trajectories, using a 10-dimensional state vector and continuous action space $\mathbf{a} = (\text{heading}, \text{speed fraction})$.

\subsubsection{Results}

\paragraph{Main model performance.} The full \pier{} agent (200 epochs) successfully navigates all three routes (Supplementary Table~\ref{stab:sim_main}). On S1 (open water), the agent completes the voyage in 16\,h with 12.3\,t fuel consumption and HF exposure of 2.63, with zero safety-shield interventions. On S2 (constrained corridor), the agent navigates the peninsula passage cleanly in 16\,h, demonstrating learned land-avoidance. On S3 (storm crossing), the agent finds a route through the high-$H_s$ region in 17\,h with only a single shield intervention.

\begin{table}[h]
\centering
\caption{\textbf{\pier{} main model performance on simulated routes} (200-epoch IQL). All three routes are completed successfully.}
\label{stab:sim_main}
\begin{tabular}{lccccc}
\toprule
\textbf{Route} & \textbf{Arrived} & \textbf{Time (h)} & \textbf{Fuel (t)} & \textbf{HF Exp.} & \textbf{Shield Int.} \\
\midrule
S1 (Open Water)       & \checkmark & 16.0 & 12.3 & 2.63 & 0 \\
S2 (Corridor)         & \checkmark & 16.0 & 16.1 & 3.56 & 0 \\
S3 (Storm)            & \checkmark & 17.0 & 16.0 & 3.73 & 1 \\
\bottomrule
\end{tabular}
\end{table}

\begin{figure}[h]
\centering
\includegraphics[width=\textwidth]{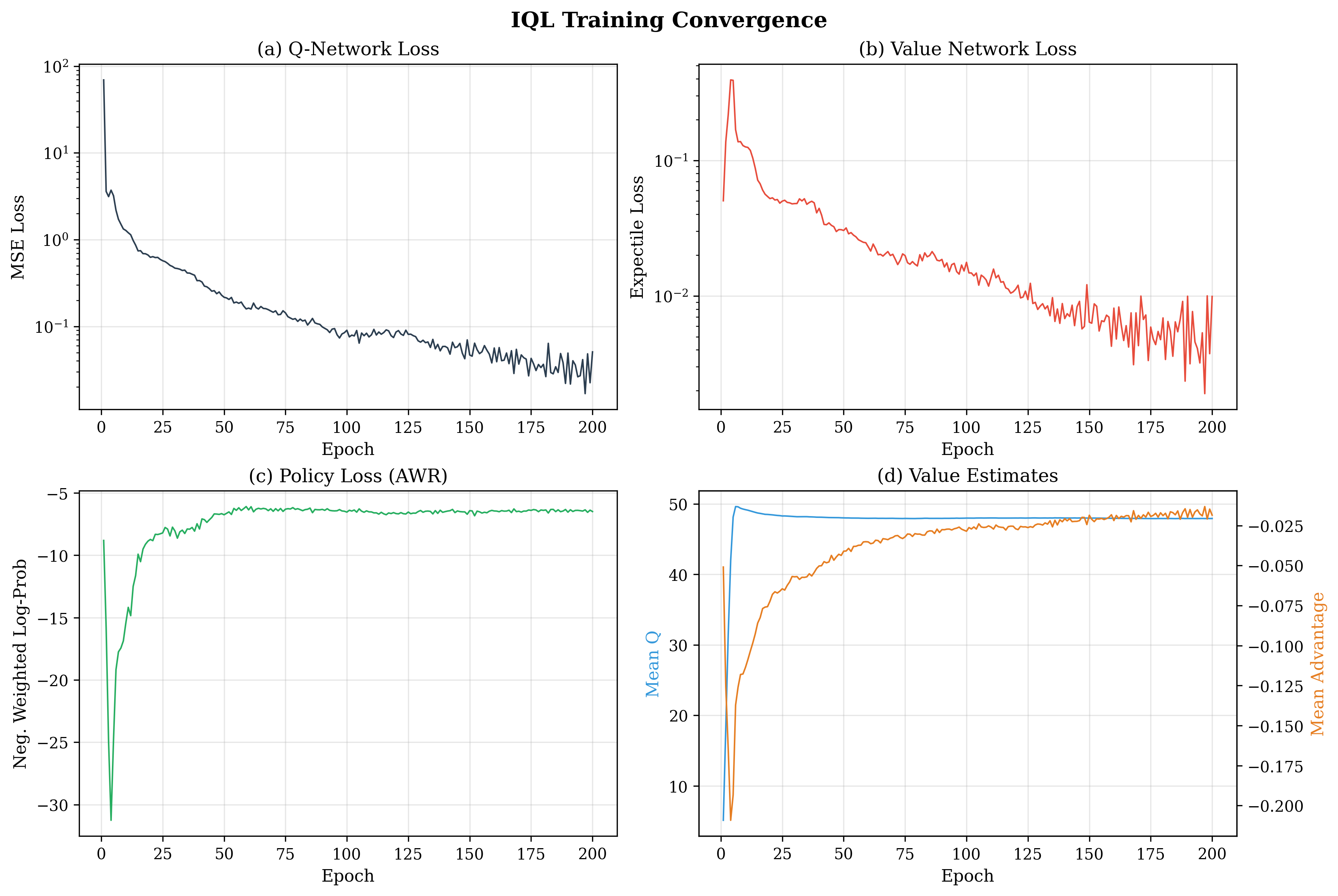}
\caption{\textbf{IQL training convergence.}
\textbf{a},~Q-network MSE loss.
\textbf{b},~Value network expectile loss.
\textbf{c},~Policy loss (advantage-weighted regression).
\textbf{d},~Mean Q-value and mean advantage. All losses plateau by epoch~150.}
\label{sfig:convergence}
\end{figure}

\paragraph{Training convergence.} IQL training dynamics show Q-network loss decreasing by approximately three orders of magnitude over 200 epochs (Supplementary Fig.~\ref{sfig:convergence}). The policy loss stabilizes by epoch~75, while mean Q-values converge to approximately~48. The mean advantage converges near $-0.03$, indicating stable value estimation.

\paragraph{Ablation study.} Five IQL variants (100 epochs each) isolate each component's contribution (Supplementary Table~\ref{stab:sim_ablation}).

\begin{table}[h]
\centering
\caption{\textbf{Ablation study on Archipelago Basin.} The full \pier{} system (200 epochs) achieves 3/3 arrivals. Each variant (100 epochs) removes one component. Values for non-arriving routes represent state at the 300-step termination limit.}
\label{stab:sim_ablation}
\setlength{\tabcolsep}{3.5pt}
\begin{tabular}{l|ccc|ccc|ccc|ccc}
\toprule
 & \multicolumn{3}{c|}{\textbf{Time (h)}} & \multicolumn{3}{c|}{\textbf{Fuel (t)}} & \multicolumn{3}{c|}{\textbf{HF Exp.}} & \multicolumn{3}{c}{\textbf{Arrived}} \\
\textbf{Config} & S1 & S2 & S3 & S1 & S2 & S3 & S1 & S2 & S3 & S1 & S2 & S3 \\
\midrule
Full \pier{}    & 16 & 16 & 17  & 12.3 & 16.1 & 16.0 & 2.63 & 3.56 & 3.73 & \checkmark & \checkmark & \checkmark \\
\midrule
Full (100ep)    & 15 & 10 & 9   & 12.0 & 8.7  & 8.0  & 2.61 & 1.34 & 1.28 & \checkmark & $\times$ & $\times$ \\
$-$Teacher      & 15 & 6  & 17  & 11.8 & 5.5  & 16.1 & 2.52 & 0.81 & 3.77 & \checkmark & $\times$ & \checkmark \\
$-$Shield       & 15 & 19 & 9   & 12.0 & 18.1 & 8.1  & 2.63 & 3.96 & 1.29 & \checkmark & \checkmark & $\times$ \\
$-$HF Risk      & 7  & 24 & 9   & 5.1  & 23.4 & 8.1  & 0.98 & 4.52 & 1.27 & $\times$ & \checkmark & $\times$ \\
$-$Physics      & 19 & 6  & 300 & 6.8  & 2.3  & 99.5 & 3.40 & 0.81 & 68.6 & \checkmark & $\times$ & $\times$ \\
\bottomrule
\end{tabular}
\end{table}

The ablation reveals clear component hierarchy:

\emph{Removing the physics model} ($-$Physics) produces the most dramatic degradation. On S3 (storm crossing), the agent wanders for the full 300-step limit, accumulating 99.5\,t of fuel and an HF exposure of 68.6 an 18$\times$ increase over the full model. Even on S1 where it arrives, the voyage takes 19\,h versus 16\,h ($+$19\%), confirming that the physics-informed speed-loss model is critical for route efficiency.

\emph{Removing the HF-risk signal} ($-$HF Risk) causes failure on both open-water routes (S1 and S3) with 291--293 shield interventions, indicating the agent repeatedly attempts to traverse high-risk regions. On S2, where it does arrive, the voyage takes 24\,h versus 16\,h ($+$50\%) with elevated HF exposure.

\emph{Removing the safety shield} ($-$Shield) allows arrival on S2 but at higher cost (19\,h versus 16\,h, HF 3.96 versus 3.56). S3 fails without the shield, as the unconstrained policy cycles in high-wave regions.

\emph{Removing teacher demonstrations} ($-$Teacher) shows the weakest signal: S1 and S3 still succeed, but S2 fails, as behavioral data lacks demonstrations of precise peninsula navigation.

\begin{figure}[h]
\centering
\includegraphics[width=\textwidth]{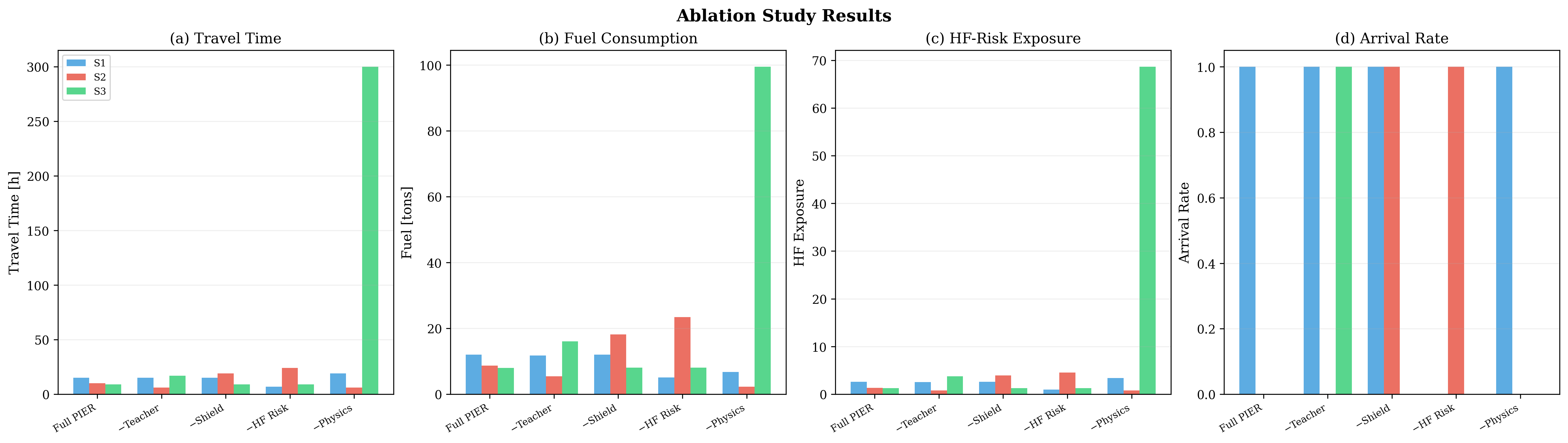}
\caption{\textbf{Ablation study on Archipelago Basin.}
\textbf{a},~Travel time: the $-$Physics variant on S3 reaches the 300\,h termination limit.
\textbf{b},~Fuel consumption: catastrophic waste on S3 without the physics model.
\textbf{c},~HF-risk exposure: $-$Physics on S3 accumulates $18\times$ the full model's exposure.
\textbf{d},~Arrival rate: only the full \pier{} system achieves 3/3 arrivals.}
\label{sfig:ablation_sim}
\end{figure}

\subsubsection{Simulated validation summary}

The Archipelago Basin validates all five \pier{} components under controlled conditions. No ablation variant matches the full model's 3/3 arrival rate with consistent travel times (15--17\,h), moderate fuel consumption (12--16\,t), and bounded HF exposure (2.6--3.7). The component hierarchy physics $>$ HF-risk $>$ shield $>$ teacher is consistent with the operational Gulf of Mexico results (main text Table~3), where the ranking is shield $>$ physics $>$ HF-risk $>$ teacher. The reversal of the top two positions reflects the difference between a simulated environment where the physics model is perfectly calibrated (making it maximally informative) and real-world operations where the safety shield compensates for model uncertainty.

\end{document}